%% file: 0_neurips_2024.tex
\documentclass{article}

\usepackage[preprint,nonatbib]{neurips_2024}

\usepackage[utf8]{inputenc} % allow utf-8 input
\usepackage[T1]{fontenc}    % use 8-bit T1 fonts
\usepackage{hyperref}       % hyperlinks
\usepackage{url}            % simple URL typesetting
\usepackage{booktabs}       % professional-quality tables
\usepackage{amsfonts}       % blackboard math symbols
\usepackage{nicefrac}       % compact symbols for 1/2, etc.
\usepackage{microtype}      % microtypography      % colors
\usepackage{amsmath}
\usepackage{multirow}
\usepackage[dvipsnames]{xcolor}
\usepackage{enumerate}
\usepackage{enumitem}
\usepackage{algorithm}
\usepackage{algpseudocode}
\usepackage{float}
\usepackage{xspace}
\usepackage{kotex}
\usepackage{setspace}
\usepackage{array}
\usepackage{lipsum}
\usepackage{ifpdf}
\usepackage[pdftex]{graphicx}

\newcommand{\ours}{Layout-and-Retouch\xspace}
\makeatletter
\newcommand*{\rom}[1]{\expandafter\@slowromancap\romannumeral #1@}
\makeatother

\title{\ours: A Dual-stage Framework for Improving Diversity in Personalized Image Generation}

\author{
    \textbf{Kangyeol Kim}\textsuperscript{\rm 1,3}\textsuperscript{*},
    \textbf{Wooseok Seo}\textsuperscript{\rm 2}\textsuperscript{*},
    \textbf{Sehyun Nam}\textsuperscript{\rm 2},
    \textbf{Bodam Kim}\textsuperscript{\rm 2},\\
    \textbf{Suhyeon Jeong}\textsuperscript{\rm 2},
    \textbf{Wonwoo Cho}\textsuperscript{\rm 1,3},
    \textbf{Jaegul Choo}\textsuperscript{\rm 1,3}\textsuperscript{\dag},
    \textbf{Youngjae Yu}\textsuperscript{\rm 2}\textsuperscript{\dag}\\
    \textsuperscript{\rm 1} KAIST AI\quad\quad
    \textsuperscript{\rm 2} Yonsei University\quad\quad
    \textsuperscript{\rm 3} Letsur Inc.\quad\quad \\
    \texttt{\{kangyeolk, wcho, jchoo\}@kaist.ac.kr} \\
    \texttt{\{justin\_seo, daniel5253, qhdamm23,}\\
    \texttt{pikachuisabird, yjy\}@yonsei.ac.kr} \\    
}

\newcommand\blfootnote[1]{%
  \begingroup
  \renewcommand\thefootnote{}\footnotetext{#1}%
  \addtocounter{footnote}{-1}%
  \endgroup
}
\begin{document}

\maketitle

\blfootnote{\textsuperscript{*} These authors contributed equally to this work.}
\blfootnote{\textsuperscript{\dag} Co-corresponding authors}

\vspace{-0.5cm}
\begin{abstract}
% Given a reference image of a subject, personalized text-to-image (P-T2I) generation aims to create new, text-guided images featuring the personalized subject, where balancing prompt fidelity and identity preservation remains a critical challenge. 
Personalized text-to-image (P-T2I) generation aims to create new, text-guided images featuring the personalized subject with a few reference images. However, balancing the trade-off relationship between prompt fidelity and identity preservation remains a crtical challenge. To address the issue, we propose a novel P-T2I method called \textbf{Layout-and-Retouch}, consisting of two stages: \emph{1) layout generation} and \emph{2) retouch}. In the first stage, our step-blended inference utilizes the inherent sample diversity of vanilla T2I models to produce diversified layout images, while also enhancing prompt fidelity. In the second stage, multi-source attention swapping integrates the context image from the first stage with the reference image, leveraging the structure from the context image and extracting visual features from the reference image. This achieves high prompt fidelity while preserving identity characteristics. Through our extensive experiments, we demonstrate that our method generates a wide variety of images with diverse layouts while maintaining the unique identity features of the personalized objects, even with challenging text prompts. This versatility highlights the potential of our framework to handle complex conditions, significantly enhancing the diversity and applicability of personalized image synthesis.
\end{abstract}

\input{1_introduction}
\input{2_related_work}
\input{4_methods}
\input{5_experiments}
\input{6_conclusion}

%%%%%%%%%%%%%%%%%%%%%%%%%%%%%%%%%%%%%%%%%%%%%%%%%%%%%%%%%%%%
% \bibliographystyle{unsrtnat}
\bibliographystyle{plain}
\bibliography{main}

\appendix

\include{99_appendix}

% Checklist 
% \input{98_checklist}

%%%%%%%%%%%%%%%%%%%%%%%%%%%%%%%%%%%%%%%%%%%%%%%%%%%%%%%%%%%%

\newpage

\end{document}

%% file: 1_introduction.tex
\section{Introduction}
% P-T2I 태스크의 의미, 중요성
Following the notable success of text-to-image (T2I) generation models, \emph{e.g.,}
Stable Diffusion~\cite{rombach2022high}, which are trained on large-scale datasets of text-image pairs, there has been increasing interest in personalized text-to-image (P-T2I) generation problems~\cite{gal2022image}.
Given a few reference images containing a specific subject, P-T2I models aim to create new, prompt-guided images that include the personalized subject.
To effectively achieve this goal, previous studies~\cite{gal2022image,ruiz2023dreambooth,Ma2023SubjectDiffusionOD} proposed to learn new personalized concepts by adjusting pre-trained T2I generation models~\cite{rombach2022high,podell2024sdxl}.
These methods have demonstrated promising and visually satisfactory results, leading to the development of versatile applications with practical potential in real-world situations.

% P-T2I의 성능적 한계, 기존 Plug-in 방법론들의 적용
%The primary criterion of evaluating the generated image from  are \textit{1) identity preservation} and \textit{2) prompt fidelity}.
When evaluating P-T2I models,
there exist two primary criteria.
1) \emph{Prompt fidelity} examine the extent to which the generated image aligns with the textual description.
2) On the other hand, \emph{identity preservation} assesses whether the appearance of a subject within the image faithfully maintains the characteristics of the personalized subject.
In essence, a trade-off relationship may exist between prompt fidelity and identity preservation~\cite{lee2024direct}, thereby P-T2I models often miss to illustrate characteristics of personalized concept when it comes to strictly following prompt guidance as well as retaining the details of visual attributes of the concept~\cite{nam2024dreammatcher,hao2023vico}.

% A plausible explanation for these failure instances can be attributed to data scarcity, specifically the limited availability of training data.
% Since a trained model relies on a given few images, it is less likely 
% 기존 Plug-in 방법론들의 한계점. -> 우리 방법론의 motivation (diversity)

\begin{figure*}[t]
\begin{center} 
\centerline{\includegraphics[width=1\linewidth]{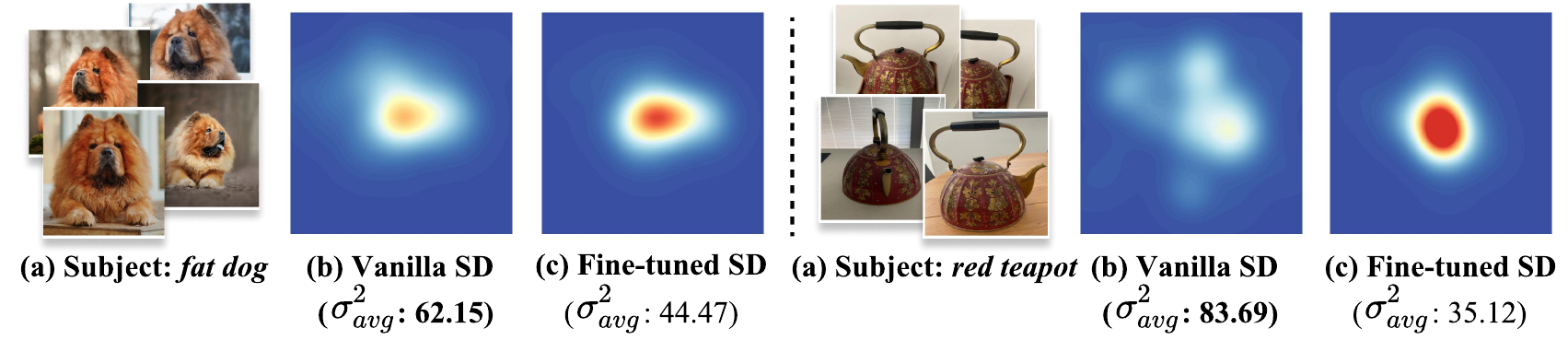}}
\vspace{-1\baselineskip}
\end{center}
\caption{
%Visualizations of center-point distributions of subjects.
%Using the ViCo~\cite{hao2023vico} evaluation prompt, we generate 10 images per prompt with both Stable Diffusion (SD)~\cite{rombach2022high} and Dreambooth~\cite{ruiz2023dreambooth} baseline trained with (a) the subject of reference images.
%For each subject, we locate the bounding box of an object using prompt-based segment-anything. 
%We then compute the 2D center-point distribution by fitting Gaussian distributions to each center point, where the mean is the center point itself and the variance is fixed, and then normalize all distributions. (b) Vanilla SD places objects across a wider range compared to (c) fine-tuned SD, indicating that the fine-tuned SD has a weaker ability to generate a diverse range of image layouts.
Visualizations of center-point distributions of subjects. Using the ViCo~\cite{hao2023vico} evaluation prompt, we generate 10 images per prompt with both Stable Diffusion (SD)~\cite{rombach2022high} and Dreambooth~\cite{ruiz2023dreambooth} trained with (a) the subject of reference images. For each subject, we locate the bounding box of an object. We then compute the 2D center-point distribution by fitting Gaussian distributions to each center point, with the mean being the center point itself and the variance fixed, and then normalize all distributions. (b) Vanilla SD places objects across a wider range compared to (c) fine-tuned SD, indicating that fine-tuned SD has a weaker ability to generate a diverse range of image layouts.
$\sigma^2_{avg}$ denotes the averaged \textit{variances} of 2D center points, meaning that the center point set is more widely dispersed with vanilla SD. 
}
\label{fig:preliminary}
\vspace{-0.6cm}
\end{figure*}

Since preserving the identity of each subject is crucial in P-T2I problems, a line of research~\cite{jia2023taming,shi2023instantbooth,su2023identity,Cao2023MasaCtrlTM,nam2024dreammatcher,zhao2023magicfusion} has focused on maintaining visual appearance during the inference phase by explicitly using reference images as an additional condition.
Although these approaches enhance the identity preservation capabilities of P-T2I models, they do not improve prompt fidelity. This is because the ability to generate contextually appropriate images is inherently limited by the intrinsic capacities of the pre-trained P-T2I model.
To investigate the inherent limitations of pre-trained P-T2I generation models in comparison with vanilla T2I models such as Stable Diffusion~\cite{rombach2022high}, we conduct exploratory experiments, the results of which are shown in Fig.~\ref{fig:preliminary}.
In fact, requiring the P-T2I model to depict the appearance of personalized concepts and synthesize non-personalized contexts is highly demanding.

% Several previous approaches~\cite{jia2023taming,shi2023instantbooth,su2023identity} .
%In this context, \textit{training-free} plug-in approaches~\cite{Cao2023MasaCtrlTM,nam2024dreammatcher,zhao2023magicfusion} have been proposed as cost-effective and effective ways to fuse visual condition.
%Technically, these approaches manipulate the keys and values of the self attention module at denoising process of U-Net, which proves to contain structures and textures information ~\cite{Cao2023MasaCtrlTM,nam2024dreammatcher}.
%Specifically, MasaCtrl~\cite{Cao2023MasaCtrlTM} is a dual-path pipeline that simultaneously synthesizes a reference and target image, while deliberately replacing target keys and values of the self-attention module with reference ones. 
%Recently, Dreammatcher~\cite{nam2024dreammatcher} proposes to adjust a replaced target value using flowmap-based semantic matching techniques to match structure correspondences between target and reference intermediate latent features.

% 우리의 구체적인 방법론 % 전반적인 이름 수정.
To overcome the limitations of the previous works and
generate personalized images from diverse and challenging prompts while retaining identity characteristics,
we propose a novel off-the-shelf P-T2I generation strategy called \textbf{\ours}.
Our core idea involves separating the generation of non-personalized parts (e.g., background and wearables) from the identity of target subjects,
thus structuring our generation process into two stages: \emph{1) layout image generation} and \emph{2) retouch}.

%, thereby reducing the burden of demanding personalized image requirements. Consequently, we have structured .

In the layout image generation stage, we primarily focus on improving prompt fidelity, irrespective of identity preservation.
Specifically, our first aim is to create a layout image that provides structural guidance for the next step.
For this purpose, we carefully design a \textbf{step-blended denoising} approach to improve the prompt fidelity of pre-trained P-T2I models by leveraging the diverse expressive capabilities of vanilla T2I generation models.
As a result, we obtain layout image features that depict an object visually resembling the target subject, occupying a specific position within the image. The remaining areas are arranged to align with the provided textual prompt.

%Technically, we propose a \textbf{step-blended denoising} approach that blends the denoising steps of a diffusion model using both vanilla T2I and P-T2I models.

In the retouch stage, the remaining goal is to precisely calibrate the target subject while preserving the background context of the layout image obtained in the previous stage.
To achieve this, we propose \textbf{multi-source attention swap}, a straightforward yet effective method that integrates context from the layout image and captures detailed visual appearance from a reference image simultaneously.
The principle of the attention swap technique is to transmit information, such as object appearance and style, from reference visual characteristics by replacing the queries, keys, and values in the target denoising steps with those derived from a reference image~\cite{Cao2023MasaCtrlTM,gu2024photoswap,gu2024swapanything,chung2023style}.

%
%Recent studies have validated the mechanism of transmitting . This has inspired us to propose 

%This two-stage framework allows us to create diversified-layout images compared to existing methods. Extensive experiments demonstrate its ability to produce diversity images and handle challenging prompts.
% Our empirical studies show that our ... , as illustrated in Fig.~\ref{fig:teaser}
%We summarize our contributions as follows:

Our two-stage method enables the creation of personalized images with diversified layouts, surpassing existing methods in variety. Extensive experiments demonstrate its capability to produce diverse images and effectively manage challenging prompts.

%% file: 2_related_work.tex
\section{Related Work}
\label{gen_inst}

\subsection{Text-Guided Image Generation and Editing}

\paragraph{T2I generation models.}

T2I generation models~\cite{xu2018attngan,nichol2021glide,ramesh2022hierarchical,rombach2022high,saharia2022photorealistic,podell2024sdxl} have been extensively studied for their ability to create realistic images from textual descriptions.
Recently, diffusion-based models~\cite{ramesh2022hierarchical,rombach2022high,saharia2022photorealistic} have demonstrated significant success in producing photo-realistic images from user-provided text, offering exceptional controllability.
However, models such as
Imagen~\cite{saharia2022photorealistic}, Stable Diffusion~\cite{rombach2022high}, and DALL-E series~\cite{ramesh2021zero,ramesh2022hierarchical,betker2023improving}, still face challenges in generating personalized images.
Specifically, these models struggle when it comes to creating images based on specific or user-defined concepts, where the identities are difficult to accurately convey through text descriptions.

\vspace{-0.5\baselineskip}
\paragraph{Text-guided image editing.}

Building on the advancements in T2I generation models, there have been active studies in editing specific images based on text inputs~\cite{li2020manigan,meng2021sdedit,hertz2022prompt,bar2022text2live,brooks2023instructpix2pix,kim2022diffusionclip,kawar2023imagic,avrahami2022blended}.
However, editing images while retaining most of the original content is challenging, as even minor modifications to the text guidance can lead to significant changes.
Therefore, early works
such as SDEdit~\cite{meng2021sdedit} and Blended-Diffusion~\cite{avrahami2022blended}
have been developed with limited image editing capabilities, \emph{e.g.}, requiring users to provide a spatial mask to specify the area for editing.
To address these limitations, Imagic~\cite{kawar2023imagic} and InstructPix2Pix~\cite{brooks2023instructpix2pix} have been designed using multi-stage training processes and multiple distinct models, respectively.
Despite their effectiveness, these editing models may not guarantee identity preservation when performing complex text-driven image transformations.

\subsection{Text-to-Image Personalization}

Addressing the limitations of T2I generation and editing models, personalization models aim to create new images using a few images of an object, while preserving the object's identity consistently.

\vspace{-0.5\baselineskip}
\paragraph{Optimization-based methods.}

Previous studies~\cite{gal2022image,ruiz2023dreambooth,kumari2023multi,han2023svdiff,voynov2023p+,hao2023vico,chen2023disenbooth,tewel2023key,han2023highly} have explored generating consistent image variations of a specified concept by embedding the concept within the textual domain of diffusion-based models, often represented by a particular token. This approach allows for the controlled generation of images that align closely with a target prompt.
Textual Inversion~\cite{gal2022image} and DreamBooth~\cite{ruiz2023dreambooth} are advanced methods for creating personalized images through diffusion-based models. Textual Inversion optimizes a textual embedding to integrate a specialized token with the target prompt, while DreamBooth extends this by adjusting all parameters of the denoising U-Net, targeting a specific token and the subject's class category. These methods enhance the precision and contextual relevance of image generation.
Research efforts have continually advanced by focusing on tuning key components, such as the cross-attention layer~\cite{kumari2023multi,han2023svdiff,tewel2023key}, or by incorporating additional adapters~\cite{hao2023vico} to enhance training efficiency and conditioning performance.
While these studies have shown promising results, they have faced limitations in preserving the appearance of subjects.

\vspace{-0.5\baselineskip}
\paragraph{Off-the-shelf methods.}

To eliminate the necessity of additional fine-tuning steps, researchers have explored plug-in T2I personalization techniques~\cite{Cao2023MasaCtrlTM,mou2023dragondiffu,duan2024tuning,lv2024pick,zhao2023magicfusion,si2023freeu,hertz2023style}.
These approaches not only enhance computational efficiency but also improve outcomes by explicitly utilizing reference images as an additional condition to capture visual appearance during the inference phase.
Technically, these methods manipulate the keys and values of the self-attention module during the denoising process of the U-Net, effectively altering structures and textures~\cite{Cao2023MasaCtrlTM,nam2024dreammatcher}. MasaCtrl~\cite{Cao2023MasaCtrlTM}, for instance, employs a dual-path pipeline that synthesizes both a reference and a target image concurrently, replacing the target's keys and values in the self-attention module with those from the reference. More recently, DreamMatcher~\cite{nam2024dreammatcher} introduced a technique that adjusts these replaced target values using flowmap-based semantic matching to ensure structural correspondences between the target and reference latent features.
However, the P-T2I model alone may present significant limitations in generating diverse layout images, often making it difficult to handle complex prompt conditions.
To overcome this issue, we propose a two-stage framework, \ours, where the vanilla T2I model is responsible for constructing an initial layout, leading to improvements in both prompt fidelity and layout diversity.

%% file: 4_methods.tex
\vspace{-1.0em}
\section{Proposed Methods}
\vspace{-0.5em}
\subsection{Preliminaries}
\vspace{-0.5em}
\paragraph{Latent diffusion model.}
%Recent advancements in text-to-image diffusion models, such as Stable Diffusion~\cite{rombach2022high}, have achieved robust and efficient image generation by performing the denoising process within the latent space using a pre-trained autoencoder.
%Specifically, a pre-trained encoder compresses an image into a latent representation $\mathbf{z}$. This is followed by a sequential process involving diffusion and iterative denoising steps using a conditional diffusion model $\epsilon_{\theta}$.
%During the denoising steps, a text condition $\mathbf{y}$ is incorporated through a cross-attention module, which guides the latent representations to align with the text condition.

Recent advancements in text-to-image diffusion models, such as Stable Diffusion~\cite{rombach2022high}, have achieved robust and efficient image generation by performing the denoising process within the latent space using a pre-trained autoencoder. Specifically, a pre-trained encoder compresses an image into a latent representation $\mathbf{z}$, followed by diffusion and iterative denoising steps using a conditional diffusion model $\epsilon_{\theta}$. During denoising, a text condition $\mathbf{y}$ is incorporated through a cross-attention module, guiding the latent representations to align with the text condition.
The training objective is formulated as
\begin{equation}
    \mathcal{L}=\mathbb{E}_{\mathbf{y}, \mathbf{z}, \epsilon, i} \left[ \Vert \epsilon - \epsilon_{\theta} (\mathbf{z}^i, i, \texttt{CLIP}(\mathbf{y}) \Vert \right],
\end{equation}
where $\epsilon \sim \mathcal{N}(0,1)$,
$\texttt{CLIP}$ denotes the text encoder of
CLIP~\cite{radford2021learning},
and $\mathbf{z}^i$ denotes a noisy version of the latent representations $\mathbf{z}$.
Also, discrete time steps $i$ are sampled from the set $\{1, ..., T\}$ uniformly at random.
For effective and robust image generation, we base our method on Stable Diffusion~\cite{rombach2022high}.
% For more in-depth description of the LDM, readers should refer to the original LDM manuscript~\cite{rombach2022high}

The latent diffusion model employs a time-conditional U-net backbone, which features multiple self-attention and cross-attention layers.
Formally, the attention mechanism can be written as
\vspace{-0.5em}
\begin{equation*}
    \texttt{Attention}(Q,K,V) = \texttt{Softmax}\left(\frac{QK^T}{\sqrt{d}}V\right),
\end{equation*}
where $Q$ represents the query encoded using latent feature maps, while $K$ and $V$ denote the key and value, respectively. These are encoded using either the latent feature maps for self-attention layers or textual embeddings for cross-attention layers.
%Recent studies~\cite{Cao2023MasaCtrlTM,nam2024dreammatcher,chung2023style,hertz2023style} has shown the benefits of using target condition images, like reference or style images, in self-attention modules.
%Furthermore, Photoswap~\cite{gu2024photoswap} has highlighted the advantage of using attention map swaps from both cross-attention and self-attention modules to maintain structure consistency with a reference image.

%Recent studies~\cite{Cao2023MasaCtrlTM,nam2024dreammatcher,chung2023style,hertz2023style,gu2024photoswap} have actively investigated the mechanism and effect of these two types of attention modules in the generation process.
%Previous observations~\cite{Cao2023MasaCtrlTM,nam2024dreammatcher,chung2023style,hertz2023style} have demonstrated the effectiveness of replacing the keys and values in the self-attention module with target condition images, such as reference or style images.
%Additionally, Photoswap~\cite{gu2024photoswap} has demonstrated the benefit of preserving a consistent structure with a reference image by swapping the attention maps obtained from both the cross-attention and self-attention modules.

\vspace{-0.5\baselineskip}
\paragraph{Pre-training personalized model.}
In general, training a personalized model requires a set of $m$ reference images $\{I^m_r\}_{m=1}^{M}$ that describe a target concept~\cite{gal2022image,ruiz2023dreambooth,chen2023disenbooth}. 
Previous approaches fine-tune either a partial~\cite{gal2022image} or the entire network~\cite{ruiz2023dreambooth} to encapsulate the target concept into the network.
After training, special text tokens such as "<*>" are used to represent the personalized information, which can be flexibly combined with additional text conditions.
Although pre-trained networks with the special text tokens have shown a remarkable capacity to convey the target object, the use of only a few training images makes them highly susceptible to overfitting~\cite{zeng2024infusion,zhou2023enhancing,chae2023instructbooth}.
% One of our observations is that the special token has a negative effect on generating prompt-aligned and diverse images (See section .).
One of our observations is that a pre-trained personalized model has weak capacities for generating images with diverse configurations as shown in Fig.~\ref{fig:preliminary}.
Throughout this paper, we denote the text condition incorporating the special text token as $\mathbf{y_p}$, while the text condition with only the special text token removed is denoted as $\mathbf{y^{-}_p}$.
Additionally, we define a neutral text condition, which includes only the concept without any specific text guidance, as $\mathbf{y}_r$ (\textit{e.g.,} a photo of <*>).

\begin{figure*}[t] %%% t: top, b: bottom, h: here
\begin{center} 
\centerline{\includegraphics[width=1\linewidth]{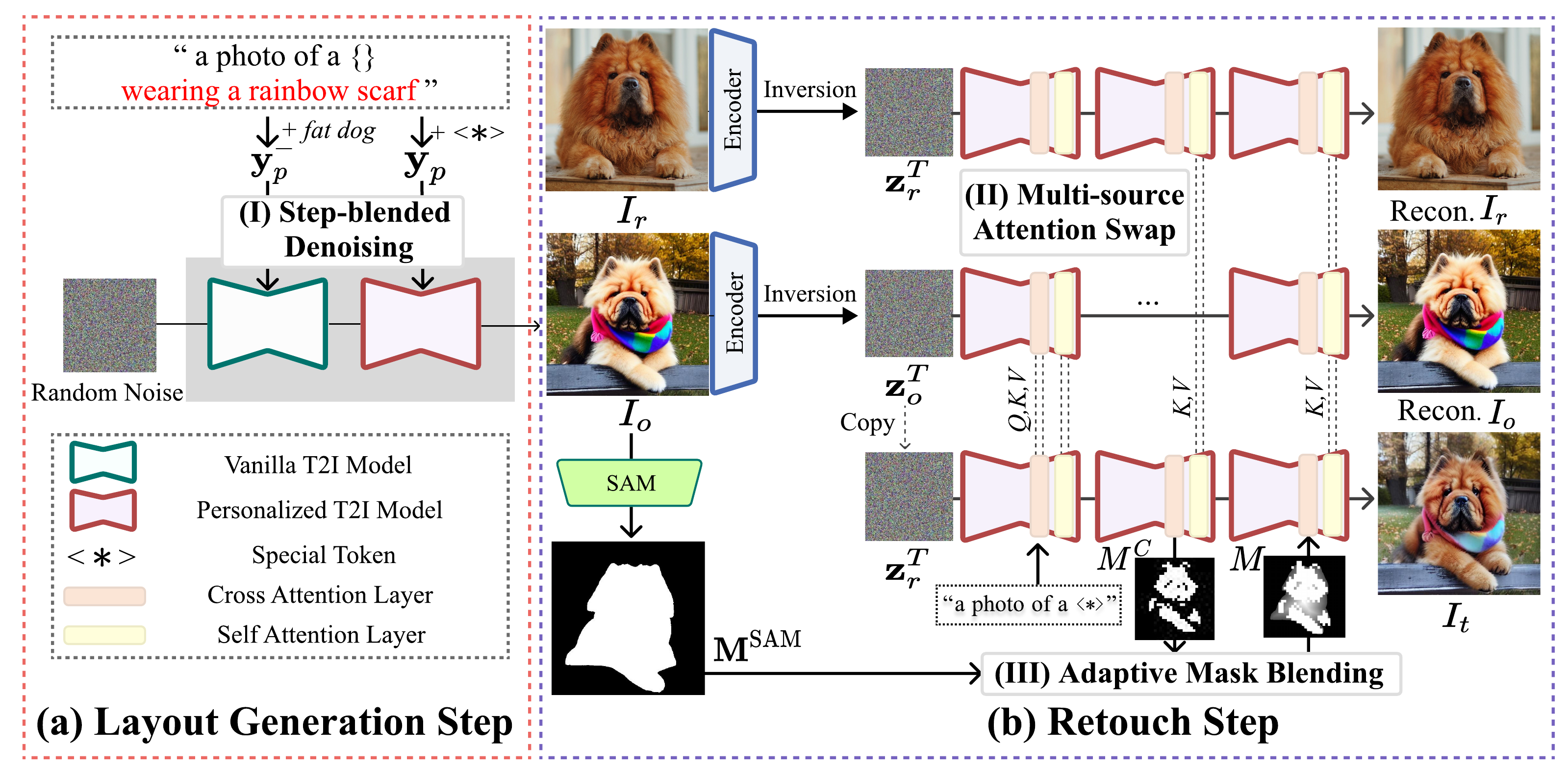}}
\vspace{-1\baselineskip}
\end{center}
\caption{
Overall pipeline of \ours:
(a) In the layout generation step, we perform (I) step-blended denoising using vanilla and personalized T2I models.
Different subject-related words (e.g., red vase and <*>) are fed to each model.
(b) The retouch step focuses on calibrating the target subject while maintaining the layout image structure. This is achieved using (II) multi-source attention swap, where intermediate variables from the attention layers of other denoising paths are used to create the target image, and (III) adaptive mask blending, which combines $\mathbf{M^{\text{SAM}}}$ and a cross-attention map to generate an accurate mask for blending feature maps in the self-attention layer.
}
\label{fig:overview}
\end{figure*}

\subsection{\ours Framework}

Given a reference image $I_r$ and the corresponding text condition $\mathbf{y}_p$, our goal is to generate a target image $I_t$ that adheres to the text condition while retaining the appearance of $I_r$.
% One solution might be to leverage $M$ with $\mathbf{y_p}$; however, we find that the pre-trained P-T2I model ends up synthesizing the restricted layout space (See ...).
One solution might be to leverage a pre-trained P-T2I model with $\mathbf{y_p}$; however, we find that the pre-trained P-T2I model tends to synthesize images within a restricted layout space (See Fig.~\ref{fig:preliminary} and Fig~\ref{fig:inception}).
We hypothesize that the model pre-trained on repetitive layouts limits its capabilities to generate diverse configurations. To address this issue, we propose the \ours framework, as depicted in Fig.~\ref{fig:overview}.

\subsubsection{Stage 1 - Layout Generation}

\paragraph{Step-blended denoising.}
% Specifically, we use the text condition $\mathbf{y^{-}_p}$ after removing the personalized special token of $\mathbf{y}_p$ ($\textit{e.g.,}$ "a photo of sks dog" "to a photo of dog").
%Our core idea is to harness the expressiveness of the vanilla Stable Diffusion (SD)~\cite{rombach2022high} by allowing it to create the layout in the initial steps.
%By delegating the task of generating the initial layout to the vanilla SD, we can broaden the range and expressiveness of the layouts.
%Specifically, we use the text condition $\mathbf{y^{-}_p}$ because the vanilla SD lacks prior knowledge of the concept.
%As demonstrated in Fig.~\ref{fig:preliminary}, the layouts generated by the vanilla SD are more diverse than those from the personalized pre-training models fine-tuned on the vanilla SD.
%By leveraging the inherent capabilities of the vanilla diffusion model to generate diverse initial layouts, We facilitate creating a broader spectrum of layout images.
%The initial layout generation is achieved within $\lambda_1$ steps, where is determined based on empirical studies.
%The detailed results and analysis of these $\lambda_1$ iterations can be found in the Appendix.

Our core idea is to utilize the expressiveness of vanilla Stable Diffusion (SD)~\cite{rombach2022high} by having it create the layout in the initial steps. Delegating the task of generating the initial layout to vanilla SD broadens the range and expressiveness of the layouts. We use the text condition $\mathbf{y^{-}_p}$ since vanilla SD lacks prior knowledge of the concept. As shown in Fig.~\ref{fig:preliminary}, layouts generated by vanilla SD are more diverse than those from pre-trained P-T2I models.
By leveraging vanilla SD's ability to generate diverse initial layouts, we facilitate the creation of a broader spectrum of layout images. The initial layout generation is achieved within $\lambda_1$ steps, determined based on empirical studies. Detailed results and analysis of these $\lambda_1$ iterations are in the Appendix.

%After generating the initial layouts within $\lambda_1$ steps, the subsequent steps adjust the initial object appearance to visually align with the target object while preserving the structure established in the early steps.
%For this purpose, we leverage $\mathbf{y_p}$ and the personalized diffusion model for remaining steps.
%Let $\epsilon_{\theta}, \epsilon^{*}_{\theta}$ represent the vanilla denoising network and the personalized denoising network, respectively, and let $\mathbf{z}^i$ denote the latent representations at each denoising $i$ step.
%Formally, computing the latent representations in step-blended denoising can be written as:

Generating the initial layouts within $\lambda_1$ steps, the subsequent steps adjust the initial object appearance to visually align with the target subject while preserving the initial structure. We use $\mathbf{y_p}$ and the personalized model for the remaining steps. Let $\epsilon_{\theta}$ and $\epsilon^{*}_{\theta}$ represent the vanilla and personalized denoising networks, respectively, and let $\mathbf{z}^i$ denote the latent representations at each denoising step $i$. Formally, computing the latent representations in step-blended denoising can be written as:

\begin{equation*}
\mathbf{z}_{i-1} = 
\begin{cases}
\texttt{Sample}(\mathbf{z}^i, \epsilon_{\theta}(\mathbf{z}^i, \mathbf{y^-_p}, i)) & \text{if } i < \lambda_1, \\
\texttt{Sample}(\mathbf{z}^i, \epsilon^{*}_{\theta}(\mathbf{z}^i, \mathbf{y_p}, i)) & \text{otherwise},
\end{cases}
\quad \text{for } i = T, T-1, \ldots, 1,
\end{equation*}

where \texttt{Sample} operation determines the next latent representations using a predicted noise.
Following the denoising steps, a decoding process is performed to generate a layout image $I_o$.

\subsubsection{Stage 2 - Retouch}

\paragraph{Multi-source attention swap.}
%Although a layout image $I_o$ captures some visual characteristics of the target object, it loses finer details of the target object.
%To compensate for the loss in detail and to enhance the detailed features, our solution leverages attention swapping techniques~\cite{Cao2023MasaCtrlTM,gu2024photoswap} by using multiple source images: $I_o$ and $I_r$.
%Specifically, intermediate variables of cross and self-attention modules such as queries, keys and values, derived from $I_o$ and $I_r$, are passed to the denoising steps of the $I_t$
%For clarity, we denote $Q^c_t,K^c_t,V^c_t$ as the queries, keys and values of the cross-attention module respectively while $Q^s_t,K^s_t,V^s_t$ represents those of self-attention module in the target denoising process. 
%In addition, throughout denoising process of the target path, we use $\mathbf{y}_r$ as a text condition to focus on modifying the visual features of the target object.

Although a layout image $I_o$ captures some visual characteristics of the target subject, it loses finer details. To enhance the details, we use attention swapping techniques~\cite{Cao2023MasaCtrlTM,gu2024photoswap} with multiple source images: $I_o$ and $I_r$.
Specifically, intermediate variables from cross and self-attention modules derived from $I_o$ and $I_r$ are passed to the denoising steps of $I_t$. We denote the queries, keys, and values of the cross-attention module as $Q^c_t, K^c_t, V^c_t$ and those of the self-attention module as $Q^s_t, K^s_t, V^s_t$ in the target denoising process. Additionally, we use $\mathbf{y}_r$ as a text condition to focus on modifying the visual features of the target object throughout the denoising process.

Algorithm~\ref{algo:multi-source-attention-swapping} describes an overall process of the Retouch stage.
%During denoising process of the target path, attention swapping to replace $Q^c_t,K^c_t,V^c_t$ or $Q^s_t,K^s_t,V^s_t$  with corresponding variables derived from the layout and context path.
%In the early steps, $I_o$ we employ $Q^c_o,K^c_o,V^c_o$ and $Q^s_o,K^s_o,V^s_o$ extracted from the layout path, integrating them into the target path during denoising process to construct the overall structure of the noisy image.
%In fact, this sharing make $I_t$ to loosely adhere to the structure of $I_o$, making it easier to follow and replicate its structure.
%In subsequent denoising steps, $K^s_r$, $V^s_r$ derived from the reference path are used to replace those in target path, with the aim of infusing the detailed appearance features from the reference object.
%Furthermore, a composited foreground mask $\mathbf{M}$ is used to combine latent representations of the target and layout path within a self-attention layer.
%We detail the creation of the composited foreground mask and the specific operations involved below.
During the target path denoising, attention swapping replaces $Q^c_t, K^c_t, V^c_t$ or $Q^s_t, K^s_t, V^s_t$ with variables from the layout and context path.
In the early steps, we use $Q^c_o, K^c_o, V^c_o$ and $Q^s_o, K^s_o, V^s_o$ from the layout path, integrating them into the target path to construct the overall structure of the noisy image. This sharing helps $I_t$ loosely adhere to $I_o$'s structure, making replication easier.
In later denoising steps, $K^s_r$ and $V^s_r$ from the reference path replace those in the target path to infuse detailed features from the reference object. Additionally, a composite foreground mask $\mathbf{M}$ combines latent representations of the target and layout path within a self-attention layer, where the details on $\mathbf{M}$ are provided below.

\begin{algorithm}
\setstretch{1}
\small
\caption{Retouch stage with multi-source attention swapping algorithm}
\label{algo:multi-source-attention-swapping}
\begin{algorithmic}[1]
\State \textbf{Inputs:} layout image $I_o$, reference image $I_r$, neutral text condition $\mathbf{y}_r$, target text condition $\mathbf{y}_p$ SAM mask $\mathbf{M}^{\text{SAM}}$, pre-trained personalized diffusion model $\epsilon^{*}_{\theta}$
\State $\mathbf{z}^T_r \leftarrow \texttt{DDIMInversion}(\texttt{Encoder}(I_r), ``\,")$ \Comment{DDIM inversion for reference image}
\State $\mathbf{z}^T_o \leftarrow \texttt{DDIMInversion}(\texttt{Encoder}(I_o), ``\,")$ \Comment{DDIM inversion for layout image}
\State $\mathbf{z}^T_t \leftarrow \mathbf{z}^T_o$ \Comment{Initialize target latent with layout latent}
\For{$i = T, T-1, \ldots, 1$}
    \State $\{{\color{blue}Q^{s}_{r},K^{s}_{r},V^{s}_{r}}\}, \epsilon_r \leftarrow \epsilon^{*}_\theta(\mathbf{z}^i_r, \mathbf{y}_r, i)$ \Comment{Reference path}
    \State $\{{\color{red}Q^{c}_{o},K^{c}_{o},V^{c}_{o}}\}, \{{\color{orange}Q^{s}_{o},K^{s}_{o},V^{s}_{o}}\}, \phi_o\leftarrow \epsilon^{*}_\theta(\mathbf{z}^i_o, \mathbf{y}_p, i)$ \Comment{Layout path}
    \State $\{{\color{teal}Q^{c}_{t},K^{c}_{t},V^{c}_{t}}\}, \{{\color{magenta}Q^{s}_{t},K^{s}_{t},V^{s}_{t}}\}\leftarrow \epsilon^{*}_\theta(\mathbf{z}^t_i, \mathbf{y}_r, i)$ \Comment{Target path}
    \If{$i > \lambda_2$}
        \State $\epsilon^* \leftarrow \epsilon^*_\theta(\mathbf{z}^t_i, \mathbf{y}_r, i, \{{\color{red}Q^{c}_{o},K^{c}_{o},V^{c}_{o}}\}, \{{\color{orange}Q^{s}_{o},K^{s}_{o},V^{s}_{o}}\})$ \Comment{Denoise w/ variables}
    \Else
        \State $\epsilon^* \leftarrow \epsilon^*_\theta(\mathbf{z}^t_i, \mathbf{y}_r, i, \{{\color{teal}Q^{c}_{t},K^{c}_{t},V^{c}_{t}}\},\{{\color{magenta} Q^{s}_{t}, } \, {\color{blue} K^{s}_{r}, V^{s}_{r}}\},\phi_o,\mathbf{M}^{\text{OURS}})$  \Comment{Denoise w/ variables and mask}        
    \EndIf
    \State $\mathbf{z}^{i-1}_{r} \leftarrow \texttt{Sample}(\mathbf{z}^i_r, \epsilon_r) $ \Comment{Sample next latent for reference image}
    \State $\mathbf{z}^{i-1}_{o} \leftarrow \texttt{Sample}(\mathbf{z}^i_o, \epsilon_o) $ \Comment{Sample next latent for layout image}    
    \State $\mathbf{z}^{i-1}_{t} \leftarrow \texttt{Sample}(\mathbf{z}^i_t, \epsilon^*) $\Comment{Sample next latent for target image}
\EndFor
\State $I_t = \text{Decoder}(\mathbf{z}^0_t)$
\State $\textbf{return}$ $I_t$
\end{algorithmic}
\end{algorithm}

\vspace{-0.5\baselineskip}
\paragraph{Adaptive mask blending.}
To enhance the visual details of the target subject, we explicitly create a foreground mask to directly blend latent representations of the layout image.
One can make
foreground masks using cross-attention maps that correlate to the object prompt tokens extracted from the decoder as in~\cite{Cao2023MasaCtrlTM},
where the maps are averaged and then thresholded to produce a binary mask $\mathbf{M}^c \in \mathbb{R}^{h \times w}$, where $h$ and $w$ represent the spatial dimensions of the latent representations.
However, we observe that $\mathbf{M}^c$ often fail to cover the entire foreground and includes noisy regions.
To mitigate this issue, we additionally harness a binary foreground mask $\mathbf{M}^{\text{SAM}} \in \mathbb{R}^{H \times W}$ where $H,W$ denote height and width respectively computed by Segment-Anything~\cite{kirillov2023segany} given a layout image. We also notice that only using $\mathbf{M}^{\text{SAM}}$ results in layout-target misalignment, since detailed appearance or locations may shift during the target generation process (details are explained in the Appendix~\ref{subsec: adaptive mask blending analysis}). Therefore, we propose an adaptive mask blending technique to mitigate these issues.

% Specifically, we first discard noisy regions of $\mathbf{M}^c$ based on $\mathbf{M}^{\text{SAM}}$ and apply multiple operations between $\mathbf{\tilde{M}}^c$ and $\mathbf{M}^{\text{SAM}}$ to combine two binary masks.

Let $\mathbf{M}[x,y]$ be the value at position $(x, y)$ and $\texttt{Resize}(\mathbf{M}^c) \in \mathbb{R}^{H \times W}$ be the resized mask from the averaged cross-attention map.
We first discard noisy regions of $\mathbf{M}^c$ based on $\mathbf{M}^{\text{SAM}}$, \emph{i.e.,}
\begin{align*}    
    % \mathbf{\tilde{M}}^c(x,y) &= \texttt{Resize}(\mathbf{M}^c)(x,y) \: \texttt{AND} \: \mathbf{M}^{\text{SAM}}(x,y), \\
    % \mathbf{M}^k(x,y) &= \texttt{OR}(\mathbf{\tilde{M}}^c(x,y), \mathbf{M}^{\text{SAM}}(x,y)), \;    
    \mathbf{M}^k[x,y] &= \texttt{OR}(\texttt{Resize}(\mathbf{M}^c)[x,y], \mathbf{M}^{\text{SAM}}[x,y]), \;
\end{align*}
where $\texttt{OR}$ represents pixel-wise operations between two binary values.
Then, we apply a distance transformation~\cite{rosenfeld1968distance} to smooth the abrupt transitions between regions with values of 1 and regions with values of 0.
Lastly, we add the area with 1s of $\mathbf{M}^c$ to explicitly insert confident regions based on the activated values of $\mathbf{M}^c$. 
Formally, this can be formulated as follows:
\begin{align*}
    \mathbf{M}[x,y] = \texttt{Normalize}(\mathcal{D}_{l2}(\mathbf{M}^k[x,y])) + \mathcal{C}(\mathbf{M}^c[x,y]),
\end{align*}
where $\mathcal{D}_{l2}$ denotes the distance transformation using $l2$-distance and $\texttt{Normalize}$ is a scaling operation to map the values to the range [0.5, 1.0].
Also a small connected set removal operation $\mathcal{C}$ performs to discard noisy pixel mask areas by removing connected sets below a specified volume threshold.
This allows us to reduce the potential negative effects caused by noisy masks.

During denoising steps, $\mathbf{M} \in \mathbb{R}^{H \times W}$ is resized to match the spatial dimension of the latent representations.
To blend the latent representations between the target and layout path, weighted summation is performed in the self-attention layer:
\begin{align*}
    \phi^{*} = \mathbf{M} \odot \phi_{t} + (1 - \mathbf{M}) \odot \phi_{o},
\end{align*}
where $\odot$ represents an element-wise multiplication, and $\phi_t$, $\phi_o$ are the output latent representations of the self-attention layer, for the target and layout path, respectively. 
The combined latent representation $\phi^{*}$ is passed to the next layer after the self-attention layer.

% To avoid noisy region to be included in $M$ we exclude disconnected regions that have no neighborhood pixels to have mask region

% accurate 하기 위해서 layout image의 마스크를 이용

% 구하는 과정

% Self-attention 내부에서의 연산

% + 자세한 설명?

%% file: 5_experiments.tex
\section{Experiments}
% Experiment setup
% Dataset
% ViCo dataset - 16 concepts - 31 prompts 
% 평가시 8장씩 3,969장의 이미지

% Implementation Details
% Baselines
% Metric:
% Image (Dino, Clip)
% Text (Clip)
% MagicFusion, FreeU, Dreamcatcher와의 정량비교
% MagicFusion, FreeU, Dreamcatcher와의 정성비교

% 비교-1 : TI, DB, CD + Ours
% 정량비교, Table-1
% -> TI, DB, CD / + Ours 비교

% Table-2: Plug-in Subject-driven T2I Synthesis
% MagicFusion
% FreeU
% Dreamcatcher
% -> Normal / Challenging
% + Challenging -> 정성결과

% Diversity 비교
% IS score
% Embedding space에서의 비교 (UMAP, t-sne)
% Analysis
% Ablation study
% Dreambooth - Importance of 레이아웃 오브젝트의 유사성 (중요)
% Attention map의 액티베이션을 가지고 비교. (중요)

% ablation (wooseok)
% step-blended denoising 
% multi source attention swapping
% mask blending 

% appendix에 넣을 것
% ours step 별로 차이점
% ca map stepwise 로 보기 

\begin{figure*}[t] %%% t: top, b: bottom, h: here
\begin{center} 
\centerline{\includegraphics[width=\textwidth]{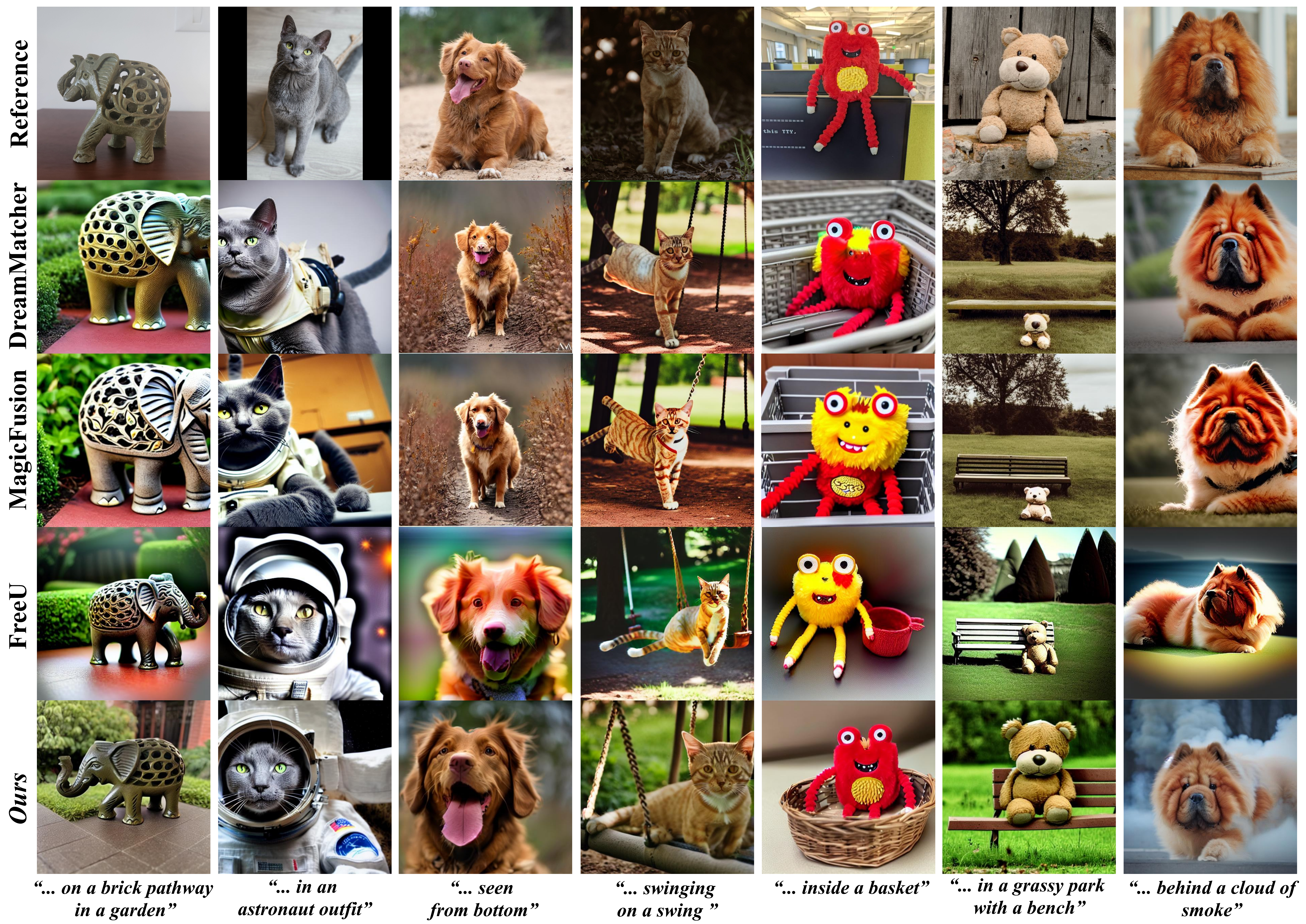}}
\end{center}
\vspace{-1\baselineskip}
\caption{
Qualitative comparisons with \textit{challenging} prompts. Our method produces images with poses and scales significantly different from the reference images compared to other methods. Additionally, it excels at generating images that accurately follow challenging prompts. It is noteworthy that our method does not produce images with identical structures, benefiting from step-blended denoising.
}
\label{fig:plugin_quali_comparison}
\end{figure*}

In this section, we describe the experimental setup (Section~\ref{subsec:setup}), including benchmark datasets, baselines, and evaluation metrics to evaluate \ours and baselines in three aspects:
\begin{enumerate}[leftmargin=*,topsep=0pt,itemsep=0ex,partopsep=1ex,parsep=1ex]
    \item \textbf{Layout Diversity:} To assess the model's ability to generate diverse layouts.
    \item \textbf{Identity Preservation:} To ensure that objects consistently maintain their detailed characteristics across images.
    \item \textbf{Prompt Fidelity:} To verify that the outputs faithfully adhere to the text guidance.
\end{enumerate}
Comparisons with baselines include both quantitative and qualitative analyses to investigate the add-on effects of \ours and to compare it with various baselines according to the criteria mentioned above (Section~\ref{subsec:comparisons}).
The qualitative analysis includes experimental results validating the superiority of \ours in generating diverse images, along with an ablation study demonstrating the contribution of each component of the proposed method.
(Section~\ref{subsec:analysis})

% substantiate our method, demonstrating the effectiveness of \ours on ? various tasks, focusing on (i) identity preservation and (ii) diversity. We ..

\vspace{-0.5em}
\subsection{Experimental Setup}
\label{subsec:setup}

\paragraph{Datasets.} 
%We adopt the image-prompt dataset proposed in ViCo~\cite{hao2023vico} for evaluation as a \textit{normal} dataset. The dataset includes a total of 16 unique concepts gathered from previous methods~\cite{gal2022image,ruiz2023dreambooth,kumari2023multi} and contains 31 distinct prompts. As for the categories, the dataset consists of 5 live animals, 6 toys, 3 household goods, 1 accessories, and 1 building, with 4-7 different images for each object. The evaluation prompt set consists of 31 prompts referenced from Dreambooth \citep{ruiz2023dreambooth}, primarily shared between living and non-living entities, with a few type-specific prompts included. In addition to this, we utilize \textit{challenging} prompt dataset proposed in DreamMatcher~\cite{nam2024dreammatcher} that includes more difficult prompts requiring large displacement or complex scene synthesis. The \textit{challenging} prompt dataset contains 24 prompts for both non-living and living entities. For evaluation, we generate 8 images per object and prompt, following the previous evaluation protocol~\cite{hao2023vico,nam2024dreammatcher}.

We adopt the image-prompt dataset proposed in ViCo~\cite{hao2023vico} for evaluation as a \textit{normal} dataset. This dataset includes 16 unique concepts from previous methods~\cite{gal2022image,ruiz2023dreambooth,kumari2023multi} and contains 31 distinct prompts. The categories include 5 live animals, 6 toys, 3 household goods, 1 accessory, and 1 building, with 4-7 images per object. The evaluation prompt set, with 31 prompts, is referenced from Dreambooth~\cite{ruiz2023dreambooth} and covers both living and non-living entities, including a few type-specific prompts.
Additionally, we use the \textit{challenging} prompt dataset proposed in DreamMatcher~\cite{nam2024dreammatcher}, which includes 24 prompts requiring large displacement or complex scene synthesis. For evaluation, we generate 8 images per object and prompt, following the evaluation protocols of previous studies~\cite{hao2023vico,nam2024dreammatcher}.

\vspace{-0.5\baselineskip}
\paragraph{Baselines.} 
Our proposed approach is compatible with \textit{any} P-T2I model without requiring additional training process. 
The primary baselines are the original P-T2I models such as Textual inversion~\cite{gal2022image}, Dreambooth~\cite{ruiz2023dreambooth}, Custom diffusion~\cite{kumari2023multi} to evaluate potential performance improvements.
Furthermore, we compare our method with other tuning-free plug-in baselines such as  FreeU~\cite{si2023freeu}, MagicFusion~\cite{zhao2023magicfusion} and DreamMatcher~\cite{nam2024dreammatcher}. 
As a backbone algorithm for evaluating plug-in baselines, we use Textual Inversion~\cite{gal2022image} and DreamBooth~\cite{ruiz2023dreambooth} as a personalized model for all baselines, including our method.

\vspace{-0.5\baselineskip}
\paragraph{Evaluation metrics.} 
% The two key aspects of evaluating P-T2I generative models are \textit{identity preservation} and \textit{prompt fidelity}. Identity preservation, which measure the quality of the synthesized image's subject details, whereas prompt fidelity means measuring the text-image alignment of the synthesized images to the given prompt. However, we tackle one more aspect of personalized text-to-image generation, which is \textbf{\textit{image diversity}}. As mentioned above, current P-T2I methods often overfit to the few user-provided images, which leads to the synthesis of reference-identical images not just the object itself, but also the backgrounds of the reference images as well. Although for a few static reference objects, generating identical results may seem ideal, since the purpose of personalized text-to-image generation lies in learning user-specific images and recreating them based on the user's purpose, copy-and-paste-like results can't be deemed preferable.
Following conventional work~\cite{gal2022image,ruiz2023dreambooth,kumari2023multi,hao2023vico}, we focus on measuring two primary aspects: (1) identity preservation and (2) prompt fidelity.
To evaluate identity preservation, we adopt $\mathbf{I_{CLIP}}$ and $\mathbf{I_{DINO}}$, which respectively employ CLIP~\cite{radford2021learning} and DINO~\cite{caron2021emerging} as backbone networks to measure subject similarity between $\{I^m_r\}_{m=1}^{M}$ and the generated images.
For evaluating prompt fidelity, we utilize the CLIP~\cite{radford2021learning} image and text encoder to compute text-image similarity score, denoted as $\mathbf{T_{CLIP}}$. 
% Also, we include ImageReward~\cite{xu2024imagereward} as a metric to assess prompt fidelity similar to MuDI~\cite{jang2024identity}.
% For clear understanding, qualitative results of justifying the use of ImageReward is provided in the Appendix~\ref{subsec:eval metric analysis}. 
Lastly, we utilize Inception Score (IS)~\cite{salimans2016improved} to measure the diversity of the generated image set.

% Finally, we first measure the diversity of the generated images with the two traditional metrics LPIPS~\cite{zhang2018unreasonable} and Inception Score~\cite{salimans2016improved}. Along with the traditional metrics, we propose our own diversity metric. 
% Also, we draw a UMAP~\cite{mcinnes2018umap} and t-SNE~\cite{van2008visualizing} to further visualize and analyze the effectiveness of our selected method. 

% TODO: Human evaluation?
% \paragraph{Human Evaluation} Since human preference for synthesized images is a very important aspect when evaluating personalized image generation, we conduct human evaluation to assess the quality of generated images between the baselines and our method. We evaluate the quality of generation with two steps, first with participants evaluating the generated images from different methods good or bad, and then battling two random methods images with the same seed to select the preferred one. If neither generated a good-quality image, participants were directed to select "neither". We also evaluate the human evaluation of diversity, where we show ? different images generated with same random seeds to the participants, to choose the best result regarding (1) image quality, (2) alignment with text, (3) diversity of the outputs. If none of the methods generated successful results, the participants were directed to select "none". Further details of human evaluation will be provided in the Appendix. 

\input{tables/comparisons_baselines_combined}
\vspace{-0.5em}
\subsection{Comparisons with Baselines}
\label{subsec:comparisons}

\paragraph{Comparisons with plug-in baselines}
We compare our approach with tuning-free plug-in baselines:  FreeU~\cite{si2023freeu}, MagicFusion~\cite{zhao2023magicfusion} and DreamMatcher~\cite{nam2024dreammatcher} on both normal and challenging prompts.
Fig.~\ref{fig:plugin_quali_comparison} shows visual comparisons of \ours and baseline outputs.
As seen, the proposed method successfully follows challenging prompt guidance while dynamically adapting the personalized object.

As shown in Table~\ref{table:baselines} (Left), the proposed method shows superior and comparable performances on both normal prompts and challenging prompts.
For \textit{normal} prompts, our method presents consistent performance increases in $\mathbf{I_{DINO}},\mathbf{I_{CLIP}}$ and $\mathbf{T_{CLIP}}$.
These improvements emphasize the effectiveness of our method in achieving both goals of personalized image generation: identity preservation and prompt fidelity. Evaluation on \textit{challenging} prompts is shown in Table~\ref{table:baselines} (Right).
Although, baseline models show comparable results on $\mathbf{T_{CLIP}}$, the $\mathbf{I_{DINO}}$ and $\mathbf{I_{CLIP}}$ scores indicate that the prompt condition for generating personalized objects is not being successfully fulfilled.
Meanwhile, our approach effectively balances the dual objectives of identity preservation and prompt fidelity in the generated images.

We conduct a user study to evaluate three aspects: (1) Identity preservation, (2) Prompt fidelity, and (3) Diversity.
As can be seen in the ranking score of Table~\ref{tab:result}, our approach is largely preferred against the other plug-in baselines. 
In the diversity aspect, we outperform DreamMatcher~\cite{nam2024dreammatcher} by a large margin, demonstrating strength in generating diverse images.
The user study setup is written in Appendix~\ref{subsec:user_study}.

% We display overall results on evaluating (1) and (2) using the average ranks, and we display the evaluation results of (3) with the win rate between \ours and DreamMatcher. 
% Note that each of the samples were randomly chosen from a large, unbiased pool. 
% Please kindly note that even when generated on the same seeds, our method does not produce images with identical structures, as the initial steps of generating layout images rely on vanilla SD. 
% We distribute the evaluation results of our study in Table~\ref{table:userstudy}, and the examples of the questions in Figure~\ref{fig:image_prompt_userstudy} and Figure~\ref{fig:diversity_userstudy}. Results show that \ours is largely preferred against the other plug-in based methodologies we used for comparison. Especially in the category of image diversity, we outperform DreamMatcher by a large margin, which scored second place in both other user studies.

% Qualitative comparisons of Fig.\ref{fig:plugin comparison} confirm...

% Furthermore, we present qualitative comparisons in Fig.~\ref{}.
% As seen in Fig.~\ref{}, our approach boosts prompt adherence compared to baselines without our method.
% Note that our method does not produce images with identical structures, as the initial steps of generating layout images rely on vanilla SD.

\input{tables/total_table}

\vspace{-0.5\baselineskip}
% \vspace{-1.5em}
\paragraph{Improving P-T2I baselines.}
We validate the effectiveness of our proposed framework by integrating it with Textual inversion~\cite{gal2022image}, Dreambooth~\cite{ruiz2023dreambooth}, Custom diffusion~\cite{kumari2023multi}.
Table~\ref{table:plug_in_table} presents the performance evaluations for identity preservation and prompt fidelity.
As shown, our approach enhances prompt fidelity across all baselines while maintaining comparable results in identity preservation.
Notably, the significant and consistent improvements in prompt fidelity demonstrate the effectiveness of our method, highlighting the importance of initial layouts in generating prompt-aligned images.
We present more qualitative comparisons with P-T2I baselines in the Appendix~\ref{subsec:additional quali}.

\subsection{Qualitative Analysis}
\label{subsec:analysis}

\paragraph{Analysis on image diversity.}

\begin{figure*}[t] %%% t: top, b: bottom, h: here
\begin{center} 
\centerline{\includegraphics[width=\textwidth]{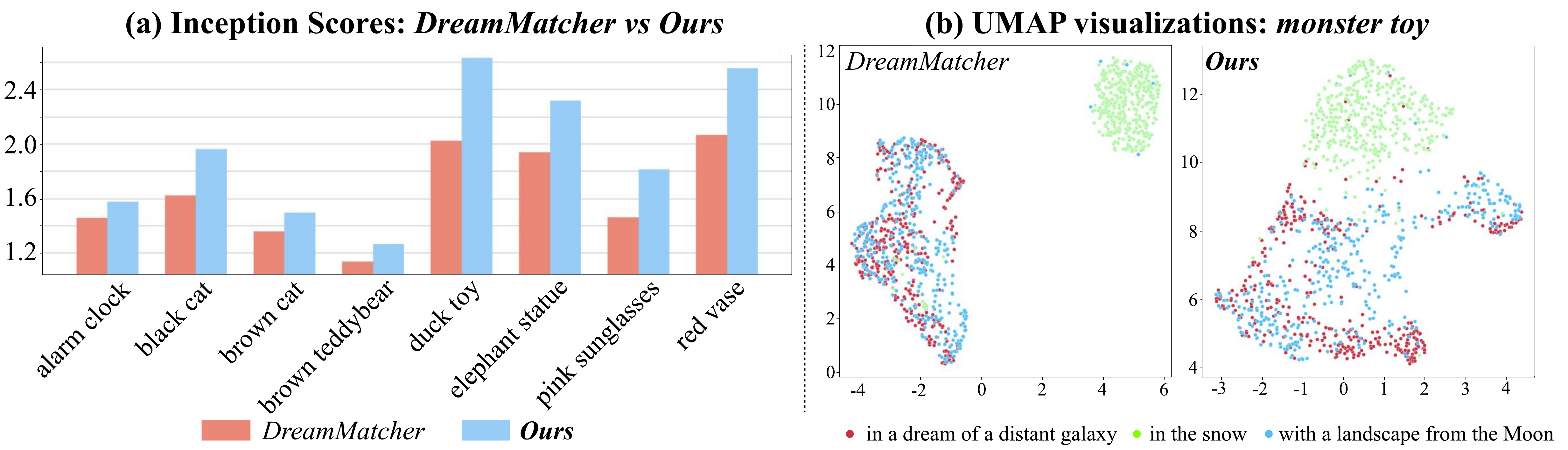}}
\vspace{-2\baselineskip}
\end{center}
\caption{
Illustrations of diversity analysis.
(a) We measure the IS score across individual objects, and verifying our method has a strong tendency to generate more diverse images.
(b) Our methodology is distributed more broadly within the same prompt.
This means that our approach is capable of generating images with diverse configurations.
}
\label{fig:inception}
\end{figure*}

One of the strengths of the proposed method lies in synthesizing images with diverse layouts.
To demonstrate this, we measure the diversity of the generated images with DreamMatcher~\cite{nam2024dreammatcher} and our approach, employing the diversity metric IS score~\cite{salimans2016improved}.
Instead of generating a small set across all prompts, we generate 400 images for 10 randomly sampled prompts out of the 31 available for each of the 8 objects with the aim of faithfully measuring the score. 
As seen in Fig.~\ref{fig:inception} (a), our method surpasses DreamMatcher~\cite{nam2024dreammatcher} in terms of IS score across all objects, indicating the superiority of ours to create images with diverse configurations.
In addition, we evaluate the diversity of the generated images by projecting three image groups of prompt using pre-trained ResNet~\cite{he2016deep}, then apply UMAP~\cite{mcinnes2018umap} to visualize the embedding space.
As shown in Fig.~\ref{fig:inception} (b) visualizes of clusters, our method obviously produces the images in broader coverage compared to DreamMatcher~\cite{nam2024dreammatcher}.
This proves the strength of \ours in generating diverse and prompt-aligned images.
The sampled prompts list is available in the Appendix~\ref{fig:diversity_prompts_list}.

% Table~\ref{table:diversity} presents % TODO.

\input{tables/ablation_table}

% \vspace{-0.5\baselineskip}
\paragraph{Ablation study.}
We study the contribution of each component by sequentially adding step-blended denoising, multi-source attention swap, and adaptive mask blending.
An ablation study is conducted based on Dreambooth baseline~\cite{ruiz2023dreambooth}.
Table~\ref{table:ablation} shows the performances of different configurations in aspects of identity preservation ($\mathbf{I_{DINO}}$, $\mathbf{I_{CLIP}}$) and prompt fidelity ($\mathbf{T_{CLIP}}$).
(\rom{2}) Building upon Dreambooth~\cite{ruiz2023dreambooth}, we introduce step-blended denoising, and demonstrate a significant enhancement in prompt fidelity.
(\rom{3}) A natural extension is to incorporate a reference image into the denoising path by utilizing the corresponding $K$ and $V$ from the self-attention layer derived from the reference image denoising process.
However, we observe a significant degradation in prompt fidelity, primarily due to the increased complexity of generating a background while accurately reflecting the visual characteristics of the reference image.
(\rom{4}) To address this, we adopt a two-stage framework with multi-source attention swapping 
This method achieves a significant performance boost in prompt fidelity compared to (\rom{3}), but it leads to a reduction in identity preservation.
(\rom{5}) In response, we employ adaptive mask blending, which enhances identity preservation and produces superior results in both identity preservation and prompt fidelity.
\vspace{-0.5em}

%% file: tables/comparisons_baselines_combined.tex
\begin{table}[!t]
\begin{center}
\centering
\fontsize{7.3}{10pt}\selectfont
\begin{tabular}{c|c|c|c||c|c|c}
\toprule
\multirow{2}{*}{\textbf{Methods}} & \multicolumn{3}{c||}{\textit{Normal Prompts}} & \multicolumn{3}{c}{\textit{Challenging Prompts}} \\ 
\cline{2-7} 
&
$\mathbf{I_{DINO}}\uparrow$ & $\mathbf{I_{CLIP}}\uparrow$ & $\mathbf{T_{CLIP}}\uparrow$ & %& $\mathbf{IR}\uparrow$ &
$\mathbf{I_{DINO}}\uparrow$ & $\mathbf{I_{CLIP}}\uparrow$ & $\mathbf{T_{CLIP}}\uparrow$ %& $\mathbf{IR}\uparrow$ 
\\
\midrule
\textbf{DreamMatcher~\cite{nam2024dreammatcher}} & \(0.576\) & \(0.781\) & \(0.278\) & \(0.650\) & \(0.824\) & \(0.310\) \\
\textbf{MagicFusion~\cite{zhao2023magicfusion}} &\(0.521\) & \(0.759\) & \(0.278\) & \(0.599\) & \(0.801\) & \(0.315\) \\ 
\textbf{FreeU~\cite{si2023freeu}} & \(0.526\) & \(0.770\) & \(0.281\) & \(0.587\) & \(0.795\) & \(0.312\) \\
\midrule
\textbf{Ours} & \(\textbf{0.600}\) & \(\textbf{0.789}\) & \(\textbf{0.297}\) & \(\textbf{0.665}\) & \(0.824\) & \(0.315\) \\
\bottomrule
\end{tabular}
\end{center}
\caption{Quantitative comparison against various plug-in based models~\cite{si2023freeu,zhao2023magicfusion,nam2024dreammatcher} on base prompts curated by ViCo~\cite{hao2023vico}. We used Textual Inversion~\cite{gal2022image} for normal prompts baseline, and we used Dreambooth~\cite{ruiz2023dreambooth} for challenging prompts baseline.}
% IR denotes as Image Reward~\cite{xu2024imagereward}.}
\label{table:baselines}
\end{table}

%% file: tables/total_table.tex
\begin{table}[htb]
  \centering
  \begin{minipage}[c]{0.6\textwidth}
    \centering
    \scriptsize
    \begin{tabular}{c|c|c|c}
    \toprule
    {\textbf{Methods}} & $\mathbf{I_{DINO}\uparrow}$ & $\mathbf{I_{CLIP}\uparrow}$ & $\mathbf{T_{CLIP}\uparrow}$ \\
    \midrule
    \textbf{Textual Inversion} & 0.527 & 0.769 & 0.277 \\
    \textit{+ Ours}            & \textbf{0.600} \textcolor{blue}{(+13.9\%)} & \textbf{0.789} \textcolor{blue}{(+2.6\%)} & \textbf{0.297} \textcolor{blue}{(+7.2\%)} \\
    \midrule
    \textbf{Dreambooth}        & 0.637 & 0.814 & 0.313 \\
    \textit{+ Ours}            & \textbf{0.655} \textcolor{blue}{(+2.8\%)} & \textbf{0.815} \textcolor{blue}{(+0.1\%)} & \textbf{0.317} \textcolor{blue}{(+1.3\%)} \\
    \midrule
    \textbf{Custom Diffusion}  & 0.658 & 0.814 & 0.315 \\
    \textit{+ Ours}            & \textbf{0.659} \textcolor{blue}{(+0.2\%)} & \textbf{0.814} \textcolor{black}{(+0.0\%)} & \textbf{0.317} \textcolor{blue}{(+0.6\%)} \\
    \bottomrule
    \end{tabular}
    \vspace{2mm}
    \caption{Add-on effects of our method. 
    Our method consistently improve baselines, validating the effectiveness of the proposed approach.
    Performance gains are marked in \textcolor{blue}{blue}.}
    \label{table:plug_in_table}
  \end{minipage}
  \begin{minipage}[c]{0.39\textwidth}
    \includegraphics[width=\textwidth]{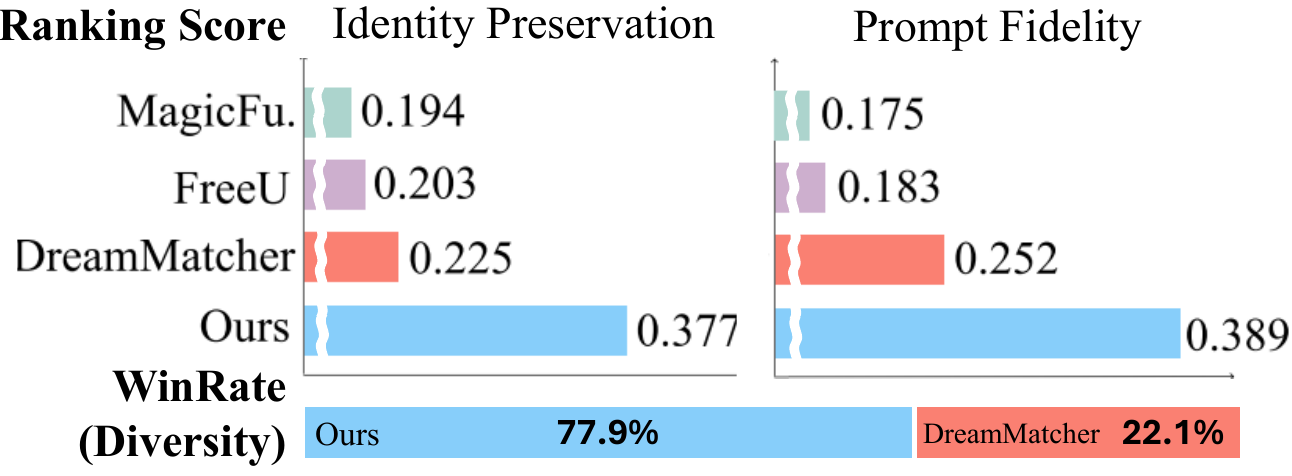}
    \caption{User study results against plug-in baselines.}
    \label{tab:result}
  \end{minipage}
  \vspace{2mm}    
\end{table}

%% file: tables/ablation_table.tex
\begin{table*}[!ht]
\begin{center}
\fontsize{9.0}{10pt}\selectfont
% \begin{tabular}{c|c|c|c|c}
% \begin{tabular}{l|l|c|c|c|c}
\begin{tabular}{l|l|c|c|c}
\toprule
& {\textbf{Configurations}} & $\mathbf{I_{DINO}}\uparrow$ & $\mathbf{I_{CLIP}}\uparrow$ & $\mathbf{T_{CLIP}}\uparrow$ \\ % & $\mathbf{IR}$ \\
\midrule
(\rom{1}) & Dreambooth~\cite{ruiz2023dreambooth} & 0.637 & 0.814 & 0.313 \\ % & 0.255 \\
\midrule
(\rom{2}) & (\rom{1}) + Step-blended denoising       & 0.620 & 0.802 & 0.319 \\ % & 0.464 \\
(\rom{3}) & (\rom{2}) + Reference attention swap        & 0.707 & 0.812 & 0.280 \\ % & -1.645 \\
(\rom{4}) & (\rom{2}) + Multi-source attention swap        & 0.615 & 0.798 & 0.320 \\ % &0.461 \\
\midrule
(\rom{5}) & (\rom{4}) + Adaptive mask blending (\textit{Ours})     & 0.655 & 0.815 & 0.317 \\ % & 0.420 \\
\bottomrule
\end{tabular}
\end{center}
\caption{
Performances of various configurations on ViCo~\cite{hao2023vico} dataset.
}
\label{table:ablation}
\end{table*}

%% file: 6_conclusion.tex
\section{Conclusion} 
\vspace{-0.5em}
% We present Layout-and-Retouch, a novel tuning-free dual-stage framework that enables a diverse and aligned generation of personalized images. 
% Our methods utilize a vanilla diffusion model in the early steps of layout generation, generating much more diverse and prompt-aligned outputs. 
% We conduct quantitative and qualitative experiments demonstrating our claims, suggesting a breakthrough in the task of personalized text-to-image generation, freeing P-T2I methodologies from their retained capabilities due to language drift in fine-tuned diffusion models. 
% We open the door for further application in our methodology, such as deploying larger-scale models or using different baselines, since our work is easily compatible.
In this paper, we present \ours, a two-stage framework for creating personalized images 
with prompt-aligned, diverse configurations.
Our motivation stems from preliminary observations on the limited layout generation capacity, which results in a weak ability to handle \textit{challenging} prompts.
To address this issue, we introduce the \textit{step-blended denoising} to diversify image layouts by leveraging the expressive power of the vanilla T2I model.
Furthermore, we propose the \textit{multi-source attention swap} and \textit{adaptive mask blending} to faithfully transfer the visual features of the reference image while preserving the layout image structure.
The combination of these modules enables our framework to effectively enrich image layouts, thereby improving its ability to handle challenging prompts.
Extensive experiments demonstrate the superiority of \ours in generating diverse prompt-aligned images, particularly in handling challenging prompts compared to baselines.

%% file: 99_appendix.tex
% Appendix에 넣기 위해 실험 중인 내용 정리
% Image Reward vs CLIPT 비교를 통한 Image Reward 사용 정당화 (완료)

% Diversity를 보여주기 위한 UMAP 추가 생성 (진행중)

% Finetuning이 디퓨전 모델의 prompt fidelity에 영향을 주는 것을 보이기 위한 tuning step에 따른 CLIPT, ImageReward curve 실험 (plotting 완료)

% 메인 컴포넌트인 vanilla > tuned model change step과 관련된 실험 / 3,5(현재),7,9,11을 비교 실험중 (실험 완료, table 화 대기 중)

% Figure 1에 그린 히트맵도 추가적으로 더 그려 넣기 (완료, 어떤 것을 넣으면 적절할지 고민 중 / 넣는게 이득인지? 이미지 보고 discussion 필요) 

% CA map inference stepwise로 visualize 해서 초반 (5) 스텝만 바닐라 모델을 사용해도 충분히 그 capability를 활용해서 이미지를 생성할 수 있다는 것을 보이기 (어느 정도 된 것은 있는데, 뭔가 썩 만족스럽지 않아서 고민중) 

% Mask Blending의 효과에 대한 퀄리 > CA map 만 썼을때 발생하는 문제, SAM만 썼을떄 발생하는 문제를 각각 이미지로 보이고 이를 우리의 blending을 썼을떄 어떻게 해결되는지 보이기 (퀄리 이미지 뽑는중, 일부 완료) 

%  ---------- 
% (If Possible) Layer 관련해서도 왜 10 ~ 15 (last two decoding layers)만 쓰는지 보이기 
% (If Possible) vanilla model에 적용한것도 점수 뽑아서 보이기 (실험 진행 중)
% Diversity 잴 때 쓰인 prompt, 그 외에도 normal prompt / challenging prompt 등등 다양한 프롬프트들 보고하기 (정리 완료, figure화 해서 추가 예정) 

%In Appendix, we first elaborate on various aspects of our study, including Broader Impact (Section ~\ref{supp_subsec:impact}), Limitations (Section ~\ref{supp_subsec:limtation}), and experimental details
In Appendix, we first discuss on broader impacts and limitations of the proposed method (Section~\ref{supp_sec:discussion}). 
Next, experimental details including implementation details are provided (Section~\ref{sec:Experimental Details}). 
Lastly, we present additional qualitative and quantitative analysis of our method(Section~\ref{sec:additionalAnalysis}). 
This includes in-depth visualizations of each component as well as further explanation about component.
Furthermore, we present a user study to compare the effectiveness and user preferences compared to baselines.

\section{Discussion}
\label{supp_sec:discussion}
\subsection{Broader Impacts.}
\label{supp_subsec:impact}
The proposed method can significantly enhance personalized image generation models, which may have various social impacts. Positively, it can facilitate the creation of more tailored and relevant content for individuals, improving user experiences across applications like social media, advertising, and digital art. Additionally, It can also assist in personalized education and training, providing custom visual aids suited to individual learning styles. However, there are potential risks, including the misuse of the technology for improving the quality of deepfakes or unauthorized digital impersonations,leading to  privacy violations and misinformation. 
Ethical concerns also arise regarding the data used for personalization, especially if it involves sensitive or personal information. Therefore, ensuring transparency, obtaining consent, and implementing robust security measures are crucial to mitigate these risks.

\subsection{Limitations.}
\label{supp_subsec:limtation}
\begin{figure*}[!ht] %%% t: top, b: bottom, h: here
\begin{center} 
\centerline{\includegraphics[width=\textwidth]{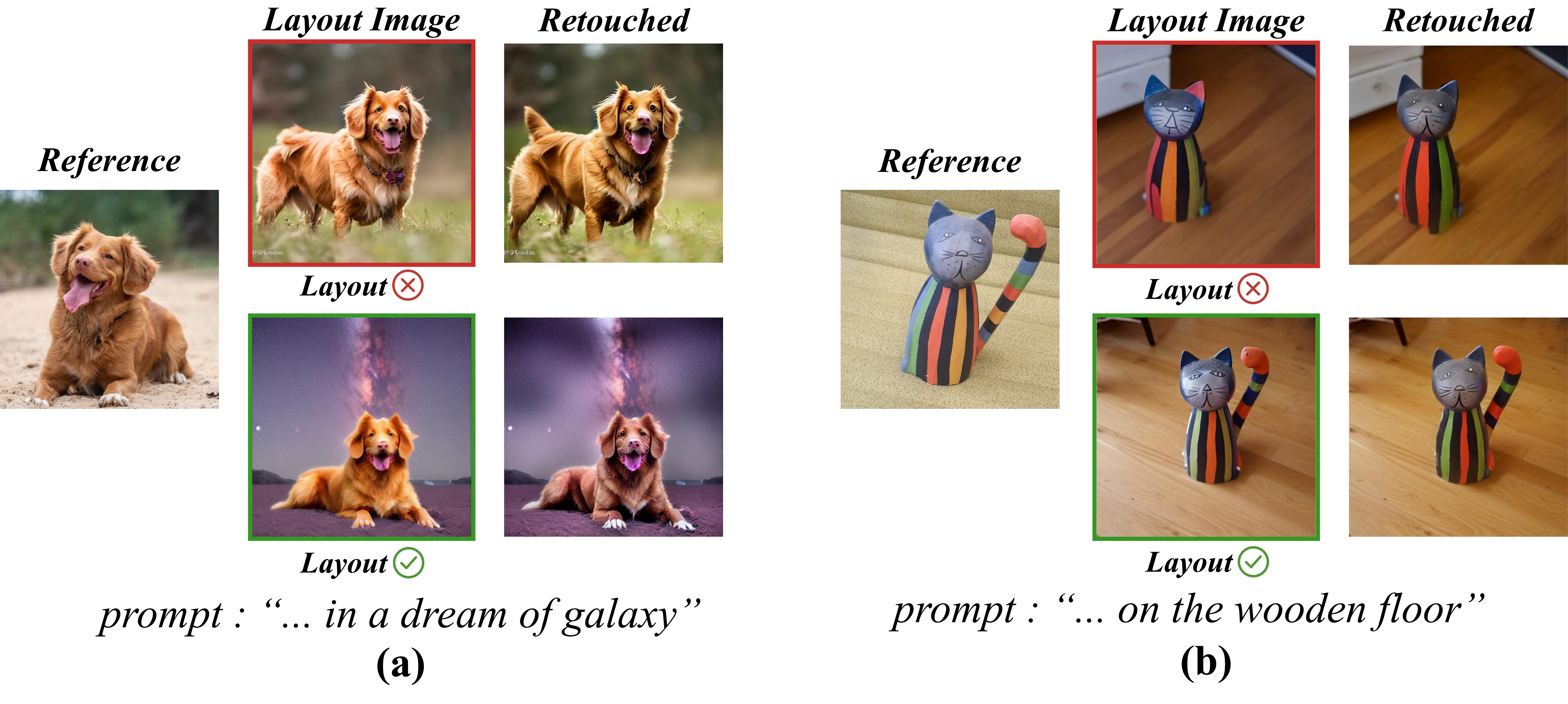}}
%\vspace{-1\baselineskip}
\end{center}
\caption{Layout Failure cases. (a) shows our generation results when layout fails to generate appropriate context, (b) shows when layout fails to faithfully generate shape of personalized concept. Since second stage of our pipeline lies on layout image, our pipeline can fail in prompts beyond the capacity of pre-trained SD backbone.}
\label{fig:limitation}
\end{figure*}

While the proposed method shows promising performances in personalized image generation, it often fails to generate an user-intended image due to the reliance on vanilla Stable Diffusion (SD) model~\cite{rombach2022high}.
Specifically, a highly complicated prompt condition that is beyond the capacity of vanilla SD can cause a failure in the layout image, subsequently leading to undesirable target image.
As seen in Fig.~\ref{fig:limitation} (a), if the layout image does not adhere to prompt condition, the final output exhibits similar issues.
Furthermore, the shape similarity between objects in layout and reference image would affect the identity preservation, as large shape discrepancies are difficult to rectify during the retouching stage.
Fig.~\ref{fig:limitation} (b) illustrates an example of generating with different layout images.
As can be seen, retaining similar shapes is beneficial for creating a personalized object characteristics in later stage.
We believe that using a more robust foundation model such as SDXL~\cite{podell2024sdxl}, to enhance prompt understanding and adherence on prompt conditions, is an effective approach to mitigating this problem.
Our future plan includes utilizing SDXL as vanilla model in step-blended denoising and developing a technique to minimize shape difference.

\section{Experimental Details}
\label{sec:Experimental Details}
\subsection{Implementation Details}
\label{subsec:Implementation Details}
We use SD 1.4 ~\cite{rombach2022high} as our baseline vanilla foundation text-to-image model, and pre-training personalized models also are conducted based on SD 1.4.
Specifically, we utilize all weights of personalized baseline models: Textual inversion~\cite{gal2022image}, Dreambooth~\cite{ruiz2023dreambooth}, Custom Diffusion~\cite{kumari2023multi} released in previous method~\cite{nam2024dreammatcher}.
In experiment, we use a DDIM sampler with a total inference time step to $T=50$.
Empirically, we set $\lambda_1=5$, $\lambda_1=3$ for normal and challenging prompt respectively and $\lambda_2=10$ for all datasets.
Additionally, the mask blend operation starts from $31^{th}$ to the last step.
For our GPU setup, we use a NVIDIA GeForce RTX 3090 GPU for all experiments, which consumes about 18GiB memory during inference.
It takes about 30 seconds for generating a single image including pre-processing. As for other parameters such as specific layers used for reference, we follow the prior works~\cite{Cao2023MasaCtrlTM,nam2024dreammatcher}.

\subsection{Prompts for Diversity Experiment}
\label{subsec: prompt dataset detail}
Since one of our core claims is the diversity of synthesized images, we demonstrate our capabilities using a few different ways. First, we visualize the center point distributions of the target subjects (see Fig.~\ref{fig:preliminary}), using all 31 prompts curated by ViCo~\cite{hao2023vico} and generating 10 images per prompt. However, when measuring diversity using classic metrics such as Inception Score~\cite{salimans2016improved} or visualizing embedding spaces~\cite{mcinnes2018umap}, it is insufficient to generate only 10 samples with each prompt. Therefore, we randomly sample 10 prompts from three categories in total, from normal prompts and challenging prompts used in DreamMatcher~\cite{nam2024dreammatcher}. These categories include large displacement, occlusion, and novel-view synthesis. We then generate 400 samples per prompt, resulting in a total of 4,000 images to evaluate Inception Scores for the objects depicted in Fig.~\ref{fig:inception}. The list of prompts is provided in Fig.~\ref{fig:diversity_prompts_list} below.

\begin{figure*}[!t] %%% t: top, b: bottom, h: here
\begin{center} 
\centerline{\includegraphics[width=\textwidth]{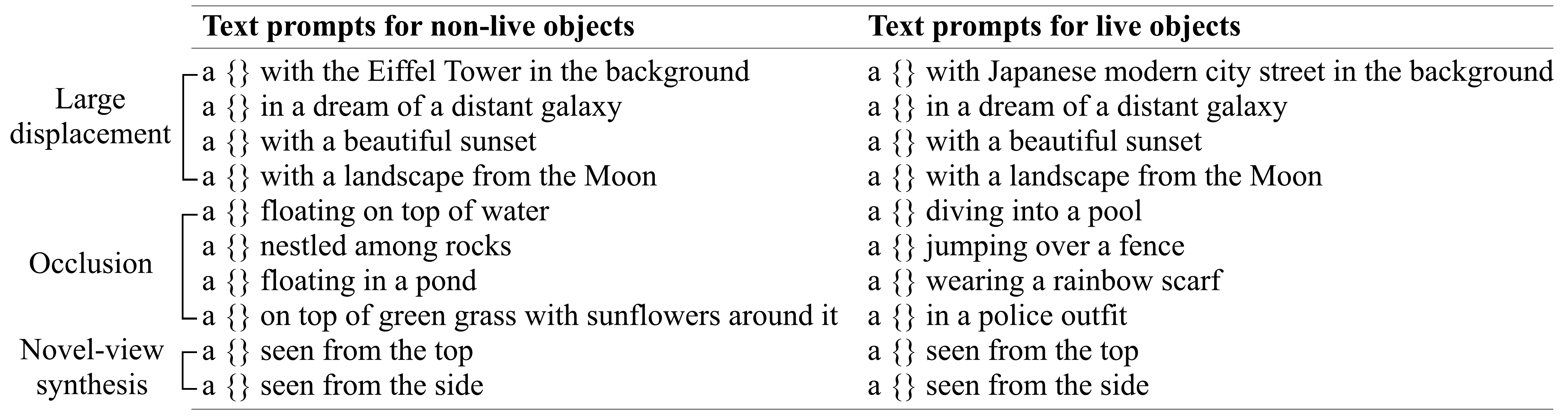}}
%\vspace{-1\baselineskip}
\end{center}
\caption{The list of 10 randomly sampled prompts that were used to generate images to compare the diversity of the output. We sampled the prompts according to the three different categories.}
\label{fig:diversity_prompts_list}
\end{figure*}

\begin{figure*}[!ht] %%% t: top, b: bottom, h: here
\begin{center} 
\centerline{\includegraphics[width=\textwidth]{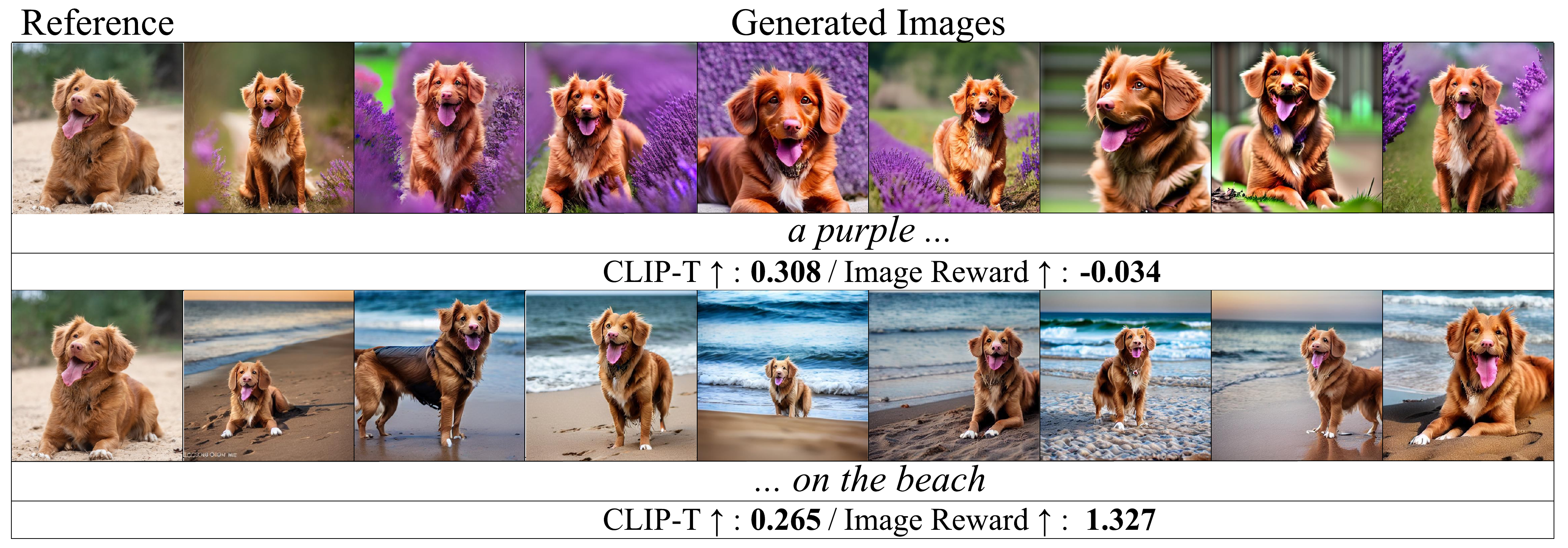}}
%\vspace{-1\baselineskip}
\end{center}
\caption{Qualitative examples of $\mathbf{T_{CLIP}}$ and ImageReward~\cite{xu2024imagereward}.
Following Instructbooth~\cite{chae2023instructbooth}, we measure both $\mathbf{T_{CLIP}}$ and ImageReward on 8 displayed images above.
Each row illustrates instances where ImageReward provide more accurate evaluations compared to $\mathbf{T_{CLIP}}$. 
Based on these observations, we additionally evaluate our method using ImageReward.}
% to justify the use of Image Reward, instead of just using CLIP-T. We show that although synthesized images don't align with the text prompts, CLIP-T still gives high scores, while Image Reward accurately measures the failed examples.}
\label{fig:CLIPT_IR1}
\end{figure*}

\subsection{Analysis on evaluation metrics}
\label{subsec:eval metric analysis}
\input{tables/imagereward_table}
% In personalized text-to-image generation task, the most common metrics DINO~\cite{caron2021emerging} similarity score and CLIP~\cite{radford2021learning} similarity scores, each denoted as $\mathbf{I_{DINO}}$, $\mathbf{I_{CLIP}}$, $\mathbf{T_{CLIP}}$. Specifically, we use ViT-S/16 DINO model as our backbone, and we use ViT-B/32 CLIP as our backbone. Although measuring cosine similarities in the two model's embedding space is a classic way to measure the capabilities of P-T2I models, recent works~\cite{chae2023instructbooth,jang2024identity,huang2024realcustom} suggest that deploying scoring functions trained on a large human
% feedback dataset such as ImageReward~\cite{xu2024imagereward} is another good way to evaluate image-text alignment. Following their claim, we analyze the qualitative results of the two metrics $\mathbf{T_{CLIP}}$ and $\mathbf{ImageReward}$. In figure~\ref{fig:CLIPT_IR1}, we show two failed generation examples from other baseline methods and a successful generation of our method. While $\mathbf{ImageReward}$ successfully measures the average scores of the generated images, we can see that $\mathbf{T_{CLIP}}$ sometimes struggles to measure the text-image alignment correctly. Therefore, we decided that using both metrics will enable us to more accurately measure the prompt fidelity of the generated images. We present out results in Table~\ref{table:image_reward}. The results show that our framework increases the prompt alignment by a large margin, which can't be demonstrated when using only $\mathbf{T_{CLIP}}$.
In main manuscript, we utilize distance-based metrics such as $\mathbf{I_{DINO}}$, $\mathbf{I_{CLIP}}$, and $\mathbf{T_{CLIP}}$ to evaluate \textit{identity preservation} and \textit{prompt fidelity}.
Specifically, ViT-S/16 DINO~\cite{caron2021emerging} and ViT-B/32 CLIP~\cite{radford2021learning} are used to extract the embeddings from the generated and input conditions such as reference images and prompts.
Although these metrics show reasonable evaluation results as previous work~\cite{Cao2023MasaCtrlTM,ruiz2023dreambooth,nam2024dreammatcher,si2023freeu,zhao2023magicfusion} demonstrated, we often face unexpected results on prompt fidelity evaluation using $\mathbf{T_{CLIP}}$.
Fig.~\ref{fig:CLIPT_IR1} illustrates inaccurate evaluation of $\mathbf{T_{CLIP}}$ on the generated images.
Specifically: (1) In Row 1: even though most images fail to comply with the prompt condition, $\mathbf{T_{CLIP}}$ remains relatively high. (2) In Row 2: despite most images successfully adhering to the prompt condition, $\mathbf{T_{CLIP}}$ is comparatively low.

To address this issue, recent work~\cite{chae2023instructbooth,jang2024identity,huang2024realcustom} suggest to employ a neural network-based evaluation model such as ImageReward~\cite{xu2024imagereward} that is trained on a large human feedback dataset.
We qualitatively validate its robustness as evaluation metric by repeatedly measuring the scores of the generated images.
Representative image sets and evaulation results are shown in Fig.~\ref{fig:CLIPT_IR1}, where ImageReward demonstrates accuracy, whereas $\mathbf{T_{CLIP}}$ as previously discussed tends to relatively inaccurate.

Upon this, we evaluate \ours using ImageReward  to determine how beneficial it is for improving prompt fidelity when plugged into the baseline.
Similar to main manuscript, we use ViCo~\cite{hao2023vico} dataset and generate 8 images per object and prompt.
Table~\ref{supp_table:plug_in_table} shows the evaluation on prompt fidelity based on both $\mathbf{T_{CLIP}}$ and ImageReward.
\ours greatly improves ImageReward across baselines, demonstrating the strength of our method to generate prompt-aligned images.

\subsection{User Study Setup}
\label{subsec:user_study}
We conduct user study with aim of evaluating three aspects: (1) \textit{Identity preservation}, (2) \textit{Prompt alignment}, (3) \textit{Diversity of outputs}. 
As baselines, we compare (1), (2) with three different prior work: MagicFusion~\cite{zhao2023magicfusion}, FreeU~\cite{si2023freeu}, DreamMatcher~\cite{nam2024dreammatcher}. Next, we evaluate (3) with DreamMatcher.

\begin{figure*}[t] %%% t: top, b: bottom, h: here
\begin{center} 
\centerline{\includegraphics[width=9cm]{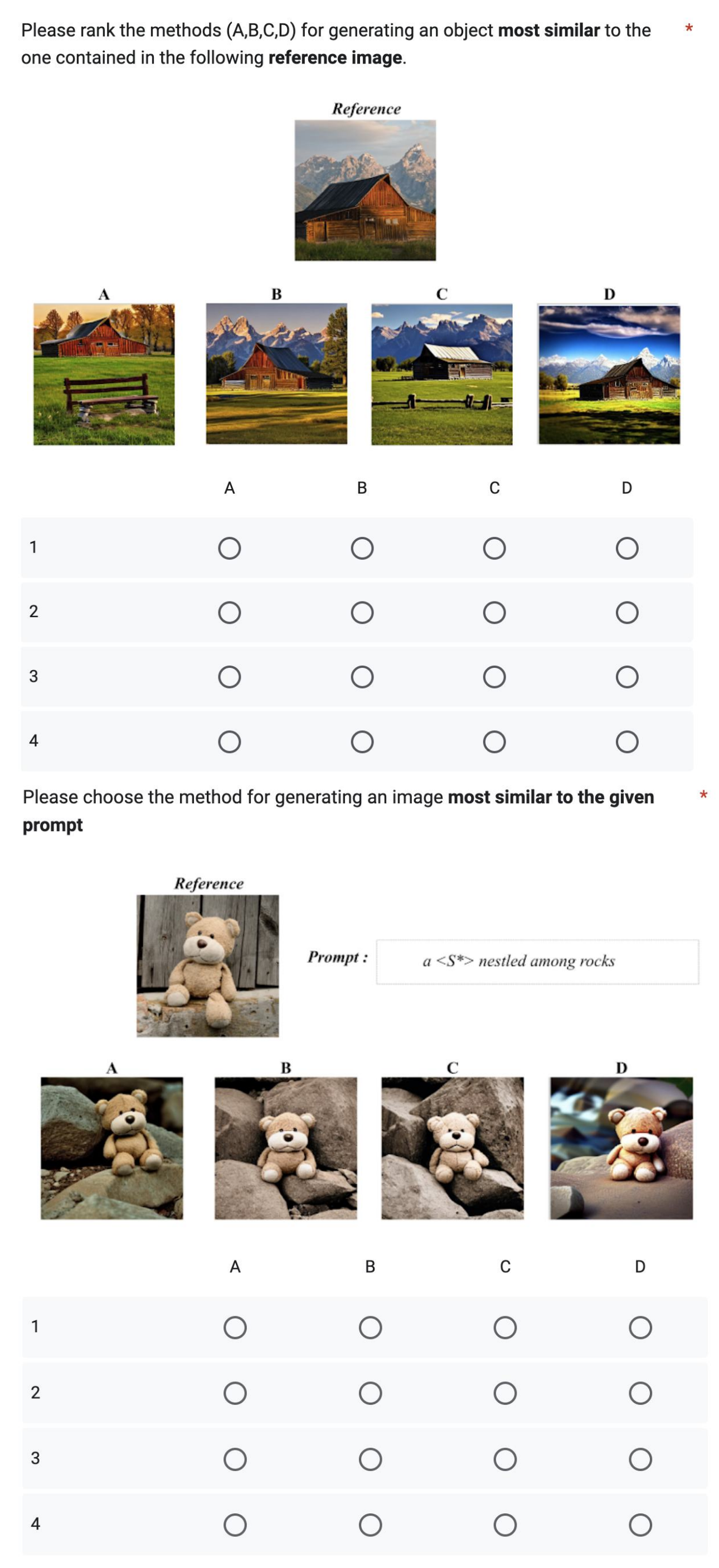}}
\vspace{-1.5\baselineskip}
\end{center}
\caption{\textbf{An example of a user study comparing \ours with prior works}: We compare Identity Preservation and Prompt Alignment with MagicFusion~\cite{zhao2023magicfusion}, FreeU~\cite{si2023freeu} and DreamMatcher~\cite{nam2024dreammatcher}. For Identity Preservation, we provide a reference image and 4 examples generated by the 4 methods with the same seed. For Prompt Alignment, we additionally provide the prompt that was used to generate the 4 images. The samples were selected randomly from a wide pool for fair comparison.}
\label{fig:image_prompt_userstudy}
\end{figure*}

\begin{figure*}[!t] %%% t: top, b: bottom, h: here
\begin{center} 
\centerline{\includegraphics[width=9cm]{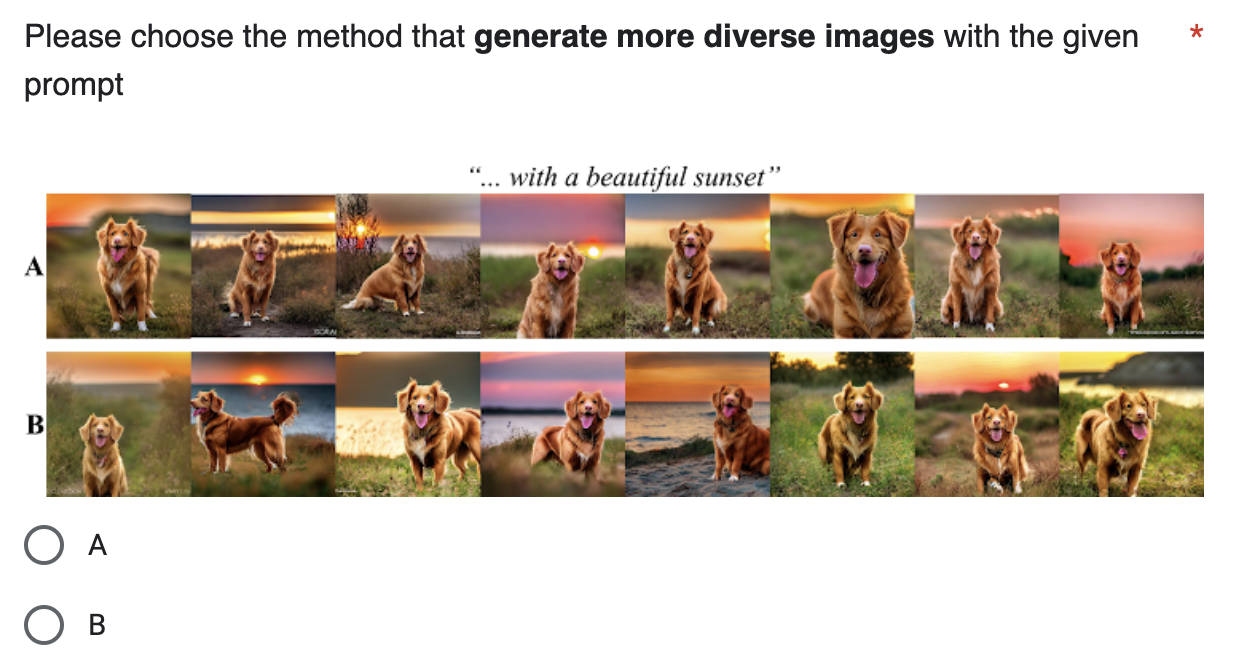}}
\vspace{-1.5\baselineskip}
\end{center}
\caption{\textbf{An example of a user study comparing \ours with prior works}: We compare Image Diversity with DreamMatcher~\cite{nam2024dreammatcher}. We provide the users with a prompt, and the 8 images generated with the same random seeds. The samples were selected randomly from a wide pool for fair comparison.}
\label{fig:diversity_userstudy}
\end{figure*}

The users were given 16 questions to compare identity preservation, 16 questions to compare prompt alignment, and 10 questions to compare diversity of generation. Two \textit{different} sets of question sets were randomly distributed in an attempt to conduct a faithful user study. 48 users answered to  42 questions, resulting in a total of 2,016 responses. 
Specifically, we design our questions as follows:
\begin{enumerate}[label=Q\arabic*:]
    \item Please rank the methods (A,B,C,D) for generating an object most similar to the one contained in the following reference image.
    \item Please choose the method for generating an image most similar to the given prompt.
    \item Please choose the method that generate more diverse images with the given prompt.
\end{enumerate}
Each question is asked to evaluate (1) \textit{Identity preservation}, (2) \textit{Prompt alignment}, (3) \textit{Diversity of outputs}, respectively.
All samples used for the user study were randomly chosen from a pool comprising tens of thousands of images generated from various prompts and object selections using ViCo~\cite{hao2023vico} dataset.
For clear understanding, we present an user interface of the user study in Fig.~\ref{fig:image_prompt_userstudy} and~\ref{fig:diversity_userstudy}

\section{Additional Analysis on \ours}
\label{sec:additionalAnalysis}

In this section, we analyze our framework \ours by providing additional experiments with aim of helping to understand the proposed method.

\subsection{Analysis on Step-blended Denoising}
\label{subsec: stepblend analysis}

\begin{figure*}[!t] %%% t: top, b: bottom, h: here
\begin{center} 
\centerline{\includegraphics[width=\textwidth]{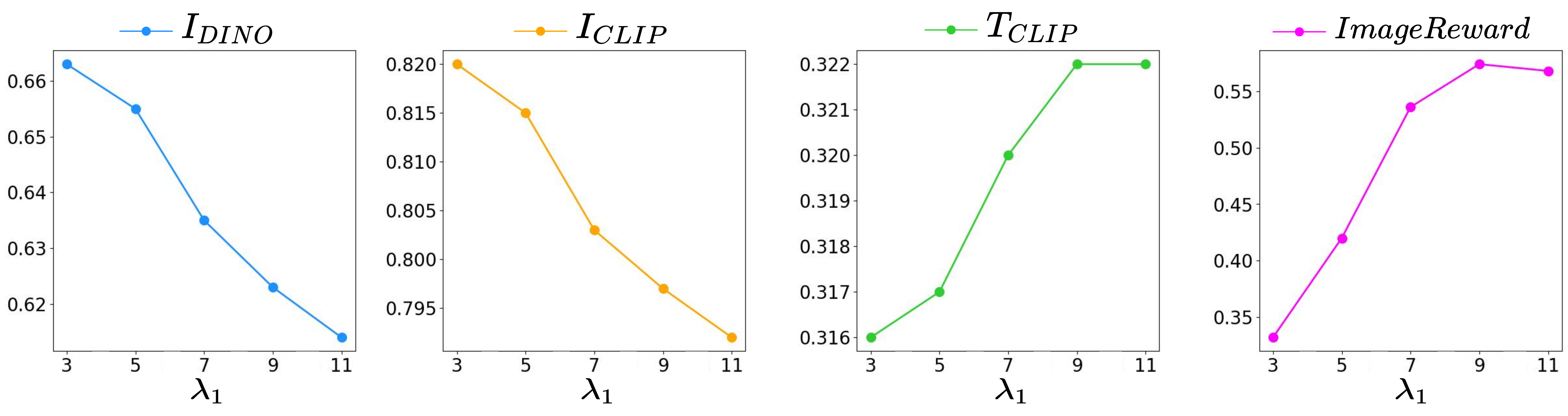}}
\vspace{-2.0\baselineskip}
\end{center}
\caption{Analysis on the effect of $\lambda_1$ of step-blended denoising. 
Note that the experiments were conducted based on 31 normal ViCo~\cite{hao2023vico} prompts using Dreambooth~\cite{ruiz2023dreambooth} baseline.}
\label{fig:changestep_analysis}
\end{figure*}

\begin{figure*}[!t] %%% t: top, b: bottom, h: here
\begin{center} 
\centerline{\includegraphics[width=\textwidth]{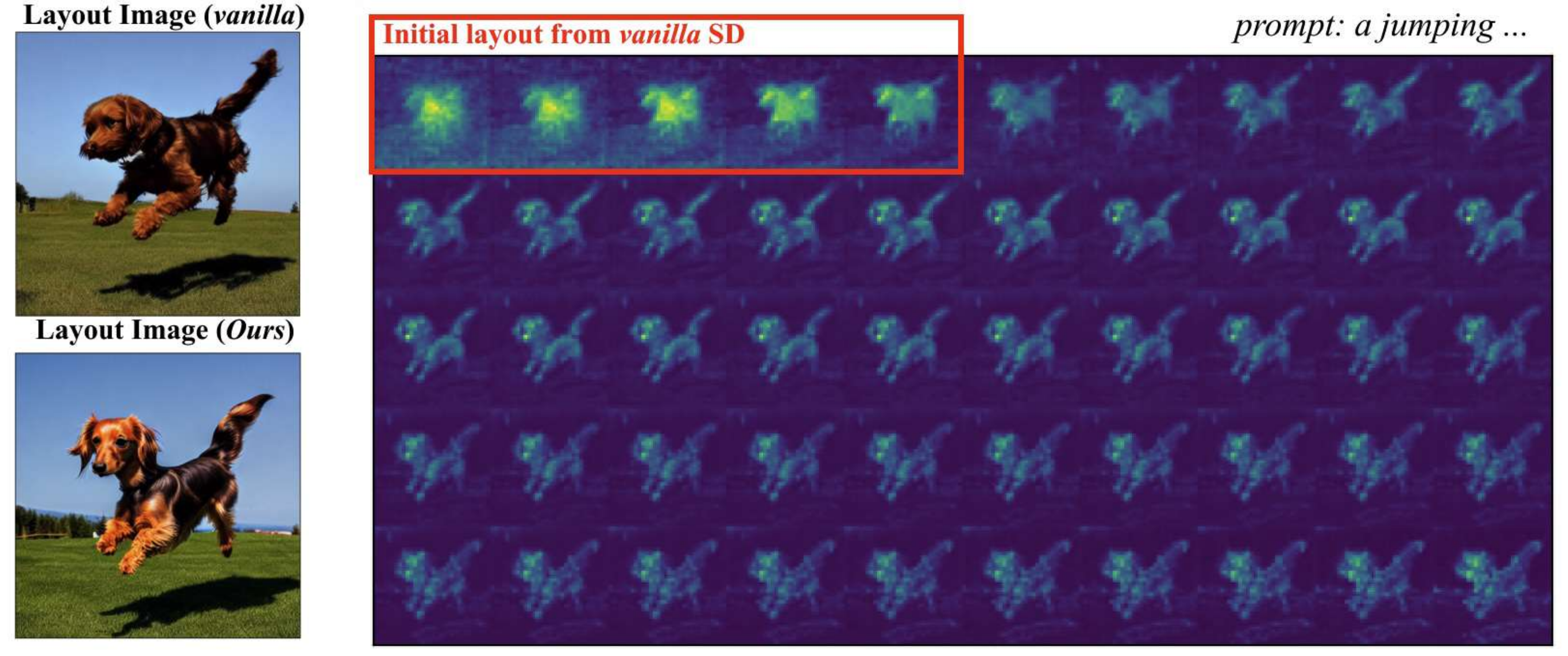}}
\vspace{-1.5\baselineskip}
\end{center}
\caption{Visualization of the cross-attention map during step-blended denoising. 
To obtain cross-attention maps, we aggregate the decoder layers of 32 x 32 resolution.}
\label{fig:camap visualize}
\end{figure*}

\paragraph{Hyperparameter sensitivity.} 
% Since our framework's main parameter is the timestep of when to change the weights of the UNet from the vanilla model to the pre-trained P-T2I model, we analyze the effects of different blend steps with quantitative experiment results. 
% As shown in Figure~\ref{fig:changestep_analysis}, we observe that earlier changes in blend steps result in higher identity preservation, while resulting in lower prompt fidelity. 
% On the other hand, when the denoising step of the vanilla model is extended, the quality of identity preservation drops rapidly while the prompt fidelity reaches an upper bound. 
% This shows that while navigating through trade-offs between identity preservation and prompt fidelity, it is crucial to limit the blend steps under 9. 
% Based on this analysis, we see that the optimal point that successfully achieves both identity preservation and prompt fidelity is step 5, since the increase rate of ImageReward~\cite{xu2024imagereward} scores is a lot higher compared to the decrease rate of identity preservation metrics, while in other change of steps, the trade-off is almost equal. 

Since synthesizing initial layout is critical for \ours, we empirically analyze the effect of the number of vanilla SD iterations in denoising steps.
As shown in Fig.~\ref{fig:changestep_analysis}, we observe that an early stop of vanilla SD denoising leads to higher identity preservation (\(\mathbf{I_{DINO}}\), \(\mathbf{I_{CLIP}}\)), while simultaneously lowering prompt fidelity (\(\mathbf{T_{CLIP}}\), \(\mathbf{ImageReward}\)). 
On the other hand, as initial layout generation steps increase, identity preservation drops rapidly while the prompt fidelity improves. 
This demonstrates a trade-off relationship between identity preservation and prompt fidelity as previous work~\cite{lee2024direct} pointed out.
Empirically, we determine our optimal $\lambda_1$ as five, where identity preservation does not dramatically drop and prompt fidelity is beyond its criterion confirmed in qualitative observation.

\paragraph{Cross-attention map visualization.} 
We visualize a cross-attention (CA) map during step-blended denoising process in Fig.~\ref{fig:camap visualize}. The two images shown in the left side are final images generated using the same seeds, using a solely vanilla denoising and step-blended denoising with $\lambda_1=5$.
We can observe that the elements of the image (\textit{e.g.,} dog pose and background) are remarkably similar, suggesting that five denoising iterations suffice to convey a diverse layout from the vanilla model to subsequent steps. 
% You can examine more precisely using the visualized cross-attention map, showing the changes in the shape of the dog through the denoising process. This visualization demonstrates that only changing the first few steps with the vanilla model is enough to generate diverse and prompt-aligned layouts, while the remaining denoising steps using the pre-trained personalized model helps the layout image to contain enough information about the user-specific objects.
Fig.~\ref{fig:camap visualize} shows detailed visualizations of cross-attention map, which illustrates the alterations in the dog's form throughout the denoising process. 
This visualization highlights that modifying only the initial steps with the vanilla model is sufficient for generating diverse and promptly aligned layouts. Subsequent denoising steps employing the pre-trained personalized model further enrich the layout image with ample information about user-specific object

\subsection{Qualitative Analysis on Mask Variants}
\label{subsec: adaptive mask blending analysis}
% In this section, we study the effects of mask blending in both quantitative and qualitative manners. Especially, we visualize the actual masks to show the effectiveness of our mask-blending method.  
%\paragraph{Quantitative Analysis on parameters of mask-blending step} As for quantitative measurements, we ablate the mask-blending steps with probing results. We evaluate the results of mask-blending step \textit{t} in every 5 steps from the 21st step to the 40th step.  We display the results in Figure ~ % 실험 진행중이므로, 자고 일어나서 그래프와 함께 작성하겠습니다.

\begin{figure*}[!t] %%% t: top, b: bottom, h: here
\begin{center} 
\centerline{\includegraphics[width=10cm]{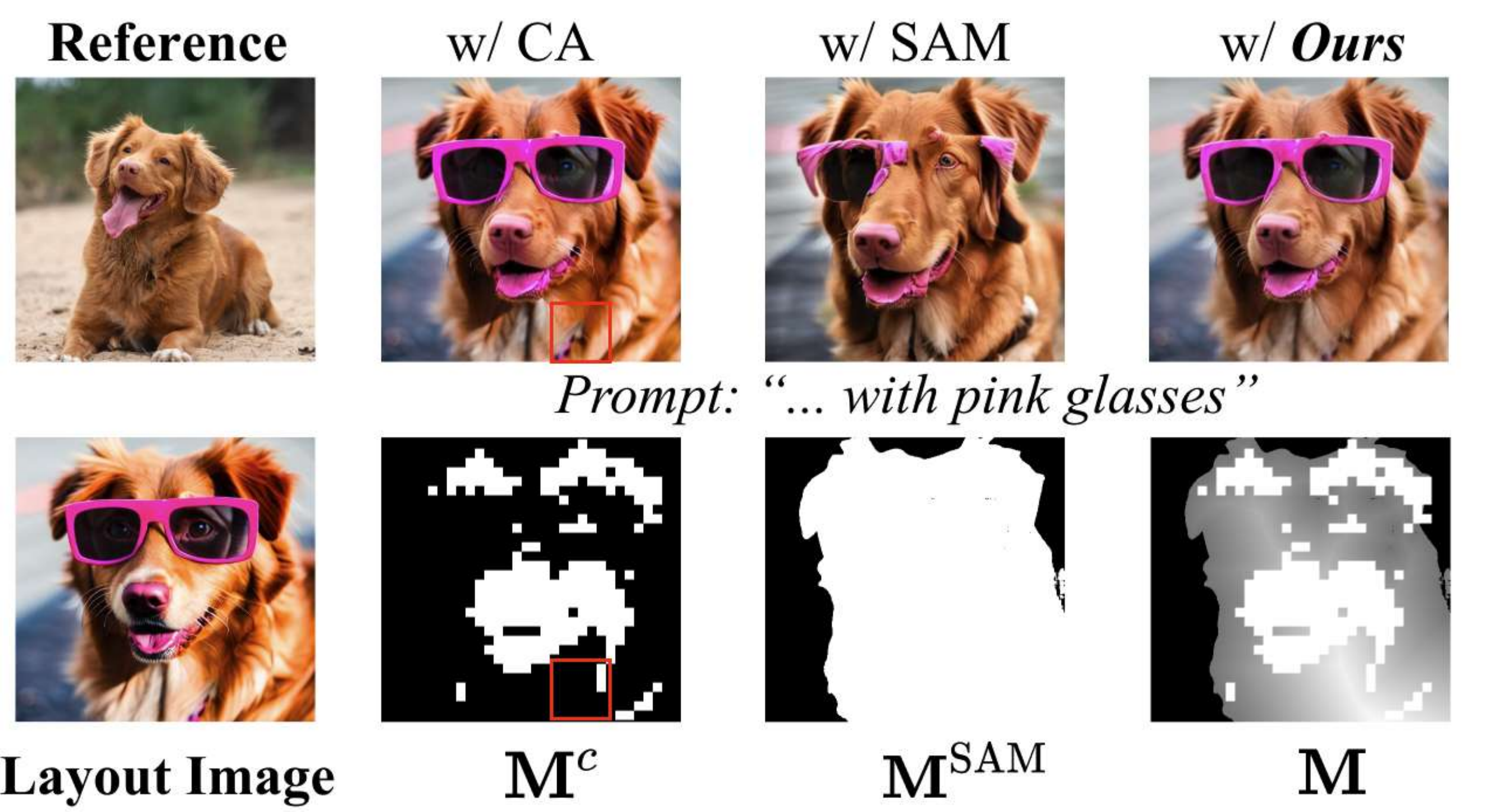}}
\vspace{-1.5\baselineskip}
\end{center}
\caption{Visualization of each foreground mask and its outputs. 
We illustrate the masks obtained from different sources and combined mask via adaptive mask blending alongside the generated image for each mask.
}
\label{fig:mask_vis}
\end{figure*}

\vspace{-0.5em}
% \paragraph{Qualitative analysis on mask variants.} 
In Fig.~\ref{fig:mask_vis}, we compare our adaptive mask blending method and the mask extracted with SAM~\cite{kirillov2023segany} and the cross-attention map after thresholding. 
In the main paper, we argued that utilizing only one of the two masks would result in either noisy region interruption or region misalignment.
As seen in Column 2 of Fig.~\ref{fig:mask_vis}, $\mathbf{M}^c$ often fail to retrieve the entire foreground object and contains noisy region, resulting in degraded performance in retaining visual details and undesirable artifacts (See \textcolor{red}{red} box).
On the other hands, Column 2 of Fig.~\ref{fig:mask_vis} shows that when using only $\mathbf{M}^{\text{SAM}}$ to define the target region, it often fails to identify important object part, impairing the necessary components of the layout image.
To address these issues, we apply adaptive mask blending, successfully overcoming them by finding a balanced mask region and strength.

\label{table:userstudy}

% \paragraph{User study guidance}
% We include the guidance for three user studies we conducted. We used short, clear instructions for users to understand clearly. Here are the three questions we asked for each category.
% \begin{enumerate}
%     \item Please rank the methods (A,B,C,D) for generating an object most similar to the one contained in the following reference image.
%     \item Please choose the method for generating an image most similar to the given prompt.
%     \item Please choose the method that generate more diverse images with the given prompt.
% \end{enumerate}
% Each question was asked to the users to evaluate the following aspects: (1) \textit{Identity Preservation}, (2) \textit{Prompt Alignment}, (3) \textit{Diversity of Outputs}. They are numerically in order. 

\subsection{Additional qualitative results}
\label{subsec:additional quali}
\begin{figure*}[!ht] %%% t: top, b: bottom, h: here
\begin{center} 
\centerline{\includegraphics[width=12cm]{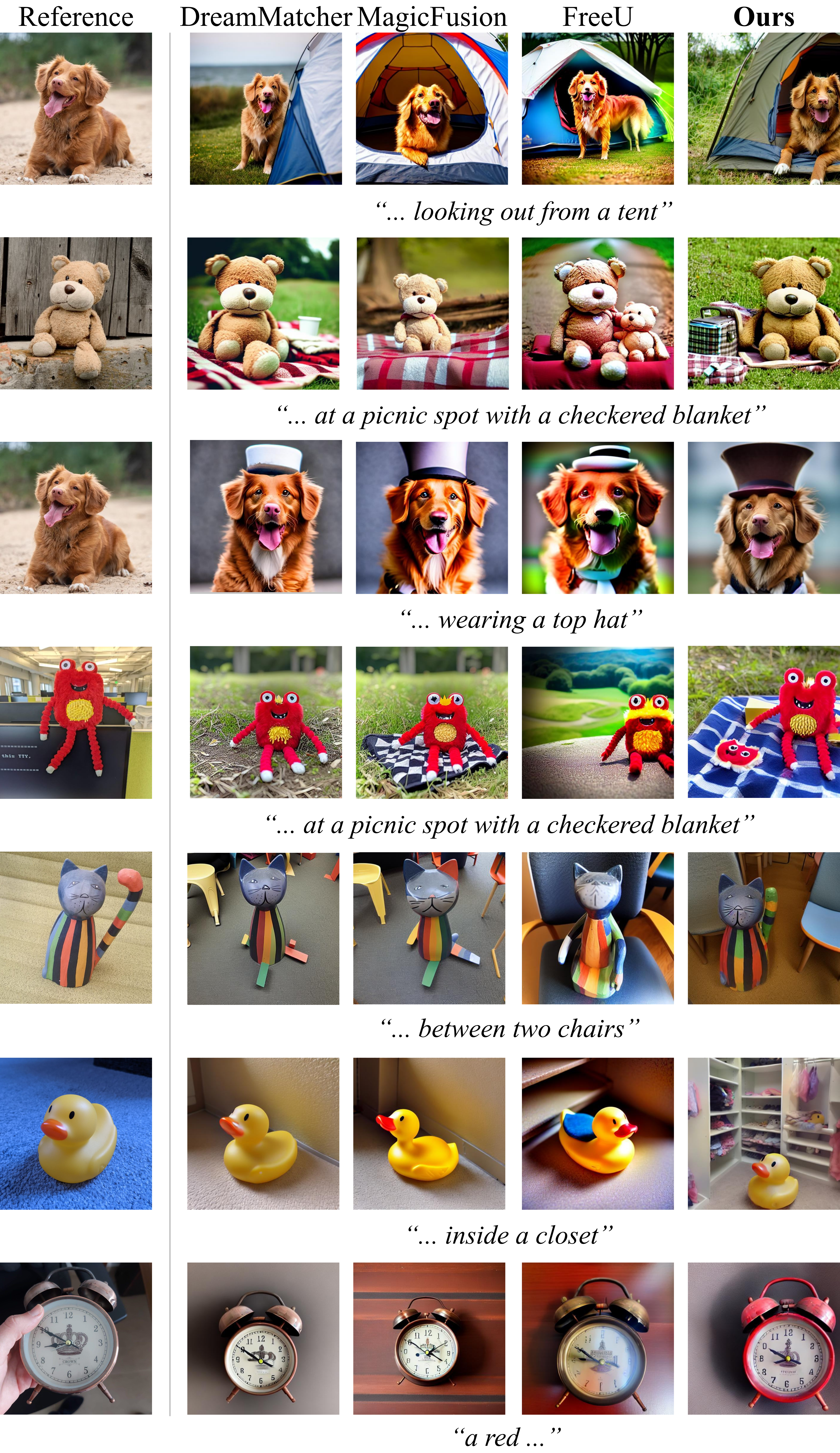}}
\vspace{-1.5\baselineskip}
\end{center}
\caption{\textbf{Qualitative comparison with plug-in based baselines~\cite{nam2024dreammatcher,zhao2023magicfusion,si2023freeu}}: We use Dreambooth~\cite{ruiz2023dreambooth} for all baselines. Note that our method does not produce images with identical structures, as the initial steps of generating layout images rely on vanilla Stable Diffusion~\cite{rombach2022high}.}
\label{fig:baseline_quali_1}
\end{figure*}

\begin{figure*}[!ht] %%% t: top, b: bottom, h: here
\begin{center} 
\centerline{\includegraphics[width=12cm]{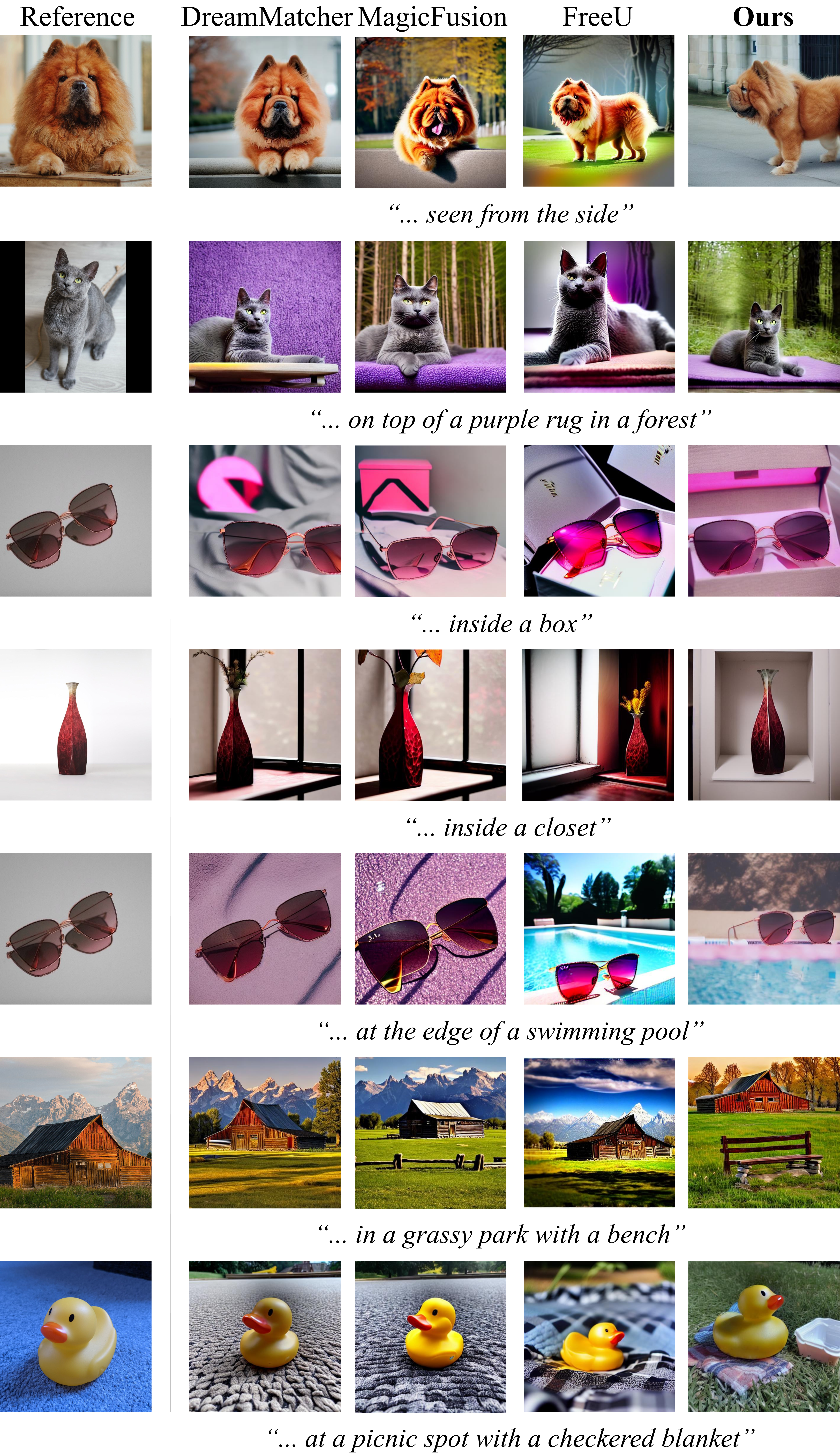}}
\vspace{-1.5\baselineskip}
\end{center}
\caption{\textbf{Qualitative comparison with plug-in based baselines~\cite{nam2024dreammatcher,zhao2023magicfusion,si2023freeu}}: We use Dreambooth~\cite{ruiz2023dreambooth} for all baselines. Note that our method does not produce images with identical structures, as the initial steps of generating layout images rely on vanilla Stable Diffusion~\cite{rombach2022high}.}
\label{fig:baseline_quali_2}
\end{figure*}

\begin{figure*}[!ht] %%% t: top, b: bottom, h: here
\begin{center} 
\centerline{\includegraphics[width=12cm]{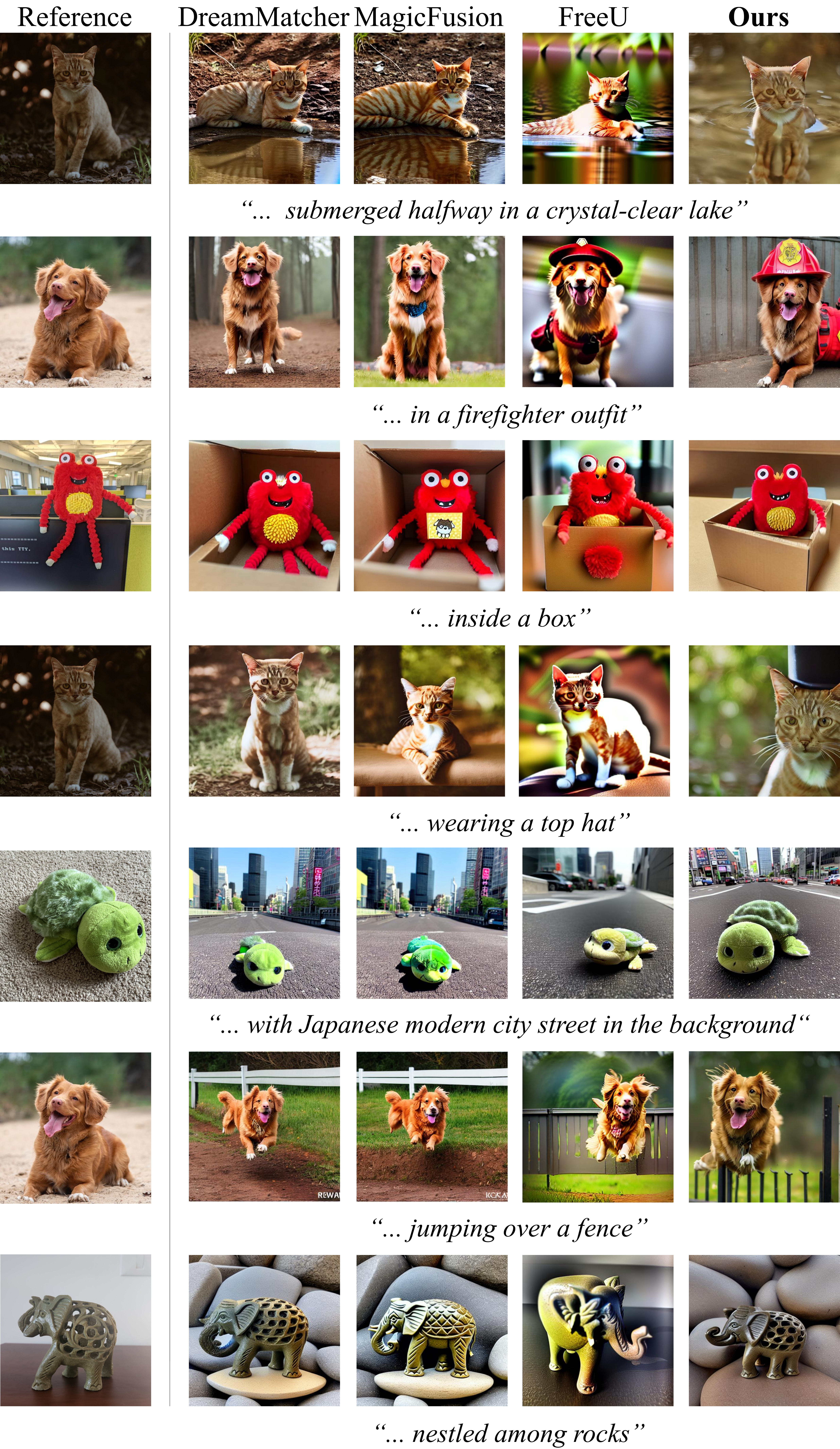}}
\vspace{-1.5\baselineskip}
\end{center}
\caption{\textbf{Qualitative comparison with plug-in based baselines~\cite{nam2024dreammatcher,zhao2023magicfusion,si2023freeu}}: We use Dreambooth~\cite{ruiz2023dreambooth} for all baselines. Note that our method does not produce images with identical structures, as the initial steps of generating layout images rely on vanilla Stable Diffusion~\cite{rombach2022high}.}
\label{fig:baseline_quali_3}
\end{figure*}

\begin{figure*}[!ht] %%% t: top, b: bottom, h: here
\begin{center} 
\centerline{\includegraphics[width=16cm]{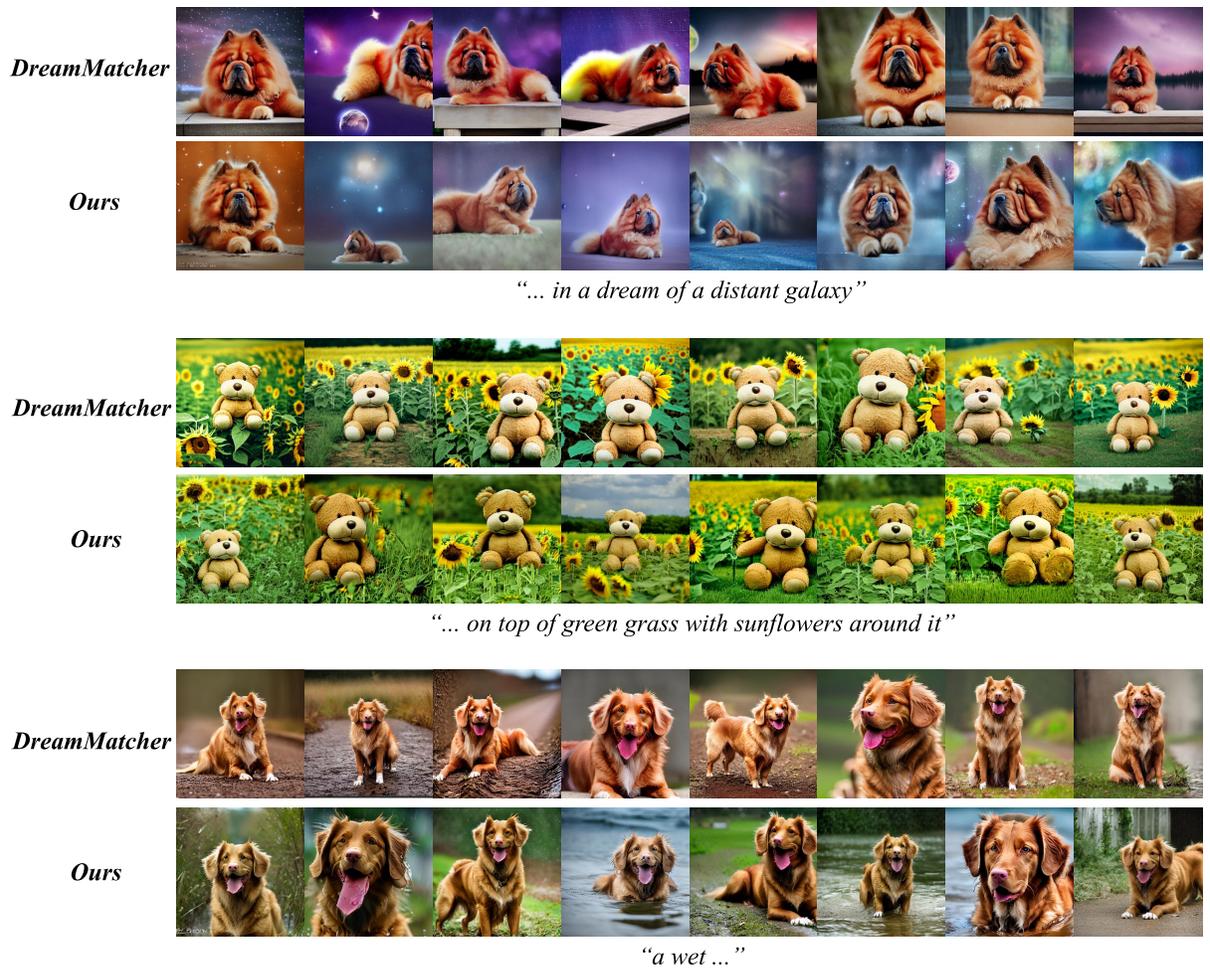}}
\vspace{-1.5\baselineskip}
\end{center}
\caption{\textbf{Diversity comparison against DreamMatcher~\cite{nam2024dreammatcher}}: We present the generated images of \ours and DreamMatcher, to highlight the capacity of the proposed approach in producing diverse images.}
\label{fig:diversity_quali}
\end{figure*}

We additionally report qualitative results of our work as follows:
\begin{enumerate}
    
    \item Additional qualitative comparisons with plug-in based baselines across various objects 
    (Fig.~\ref{fig:baseline_quali_1},  ~\ref{fig:baseline_quali_2}, and~\ref{fig:baseline_quali_3}.
    \item Additional qualitative comparisons by adding the proposed method on baselines 
    (Fig.~\ref{fig:plugin_quali_1}, ~\ref{fig:plugin_quali_2}, and~\ref{fig:plugin_quali_3}).
    \item Additional comparison with DreamMatcher on image diversity (Fig.~\ref{fig:diversity_quali})
\end{enumerate}

\begin{figure*}[!ht] %%% t: top, b: bottom, h: here
\begin{center} 
\centerline{\includegraphics[width=16cm]{figures/comparision1.pdf}}
\vspace{-1.5\baselineskip}
\end{center}
\caption{\textbf{Qualitative comparison with baselines}: We compare \ours with the three baseline methods~\cite{gal2022image,ruiz2023dreambooth,kumari2023multi} to show the effectiveness of our work. Note that our method does not produce images with identical structures, as the initial steps of generating layout images rely on vanilla Stable Diffusion~\cite{rombach2022high}.}
\label{fig:plugin_quali_1}
\end{figure*}

\begin{figure*}[!ht] %%% t: top, b: bottom, h: here
\begin{center} 
\centerline{\includegraphics[width=16cm]{figures/comparision2.pdf}}
\vspace{-1.5\baselineskip}
\end{center}
\caption{\textbf{Qualitative comparison with baselines}: We compare \ours with the three baseline methods~\cite{gal2022image,ruiz2023dreambooth,kumari2023multi} to show the effectiveness of our work. Note that our method does not produce images with identical structures, as the initial steps of generating layout images rely on vanilla Stable Diffusion~\cite{rombach2022high}.}
\label{fig:plugin_quali_2}
\end{figure*}

\begin{figure*}[!ht] %%% t: top, b: bottom, h: here
\begin{center} 
\centerline{\includegraphics[width=16cm]{figures/comparision3.pdf}}
\vspace{-1.5\baselineskip}
\end{center}
\caption{\textbf{Qualitative comparison with baselines}: We compare \ours with the three baseline methods~\cite{gal2022image,ruiz2023dreambooth,kumari2023multi} to show the effectiveness of our work. Note that our method does not produce images with identical structures, as the initial steps of generating layout images rely on vanilla Stable Diffusion~\cite{rombach2022high}.}
\label{fig:plugin_quali_3}
\end{figure*}

%% file: tables/imagereward_table.tex
\begin{table*}[!ht]
\begin{center}
\fontsize{8.5}{10pt}\selectfont
\begin{tabular}{c|c|c}
\toprule
{\textbf{Methods}} & $\mathbf{T_{CLIP}}$ & $\mathbf{ImageReward}$ \\
\midrule
\textbf{Textual Inversion} & 0.277 & -0.975 \\
\textit{+ Ours}            & \textbf{0.297} \textcolor{blue}{(+7.2\%)} & \textbf{-0.437} \textcolor{blue}{\textbf{(+55.2\%)}} \\
\midrule
\textbf{Dreambooth}        & 0.313 & 0.255 \\
\textit{+ Ours}            & \textbf{0.317} \textcolor{blue}{(+1.3\%)} & \textbf{0.420} \textcolor{blue}{\textbf{(+64.7\%)}} \\
\midrule
\textbf{Custom Diffusion}  & 0.315 & 0.257 \\
\textit{+ Ours}            & \textbf{0.317} \textcolor{blue}{(+0.6\%)} & \textbf{0.288} \textcolor{blue}{\textbf{(+12.1\%)}} \\
\bottomrule
\end{tabular}
\label{table:image_reward}
\end{center}
\caption{
The quantitative experimental results of evaluating the effect of our methodology in prompt fidelity, compared to the baseline optimization-based methods. 
We follow the dataset gathered by ViCo~\cite{hao2023vico}. 
Performance gains are indicated in blue.}
\label{supp_table:plug_in_table}
\end{table*}